\documentclass[conference]{IEEEtran}
\usepackage{times}

\usepackage[numbers]{natbib}
\usepackage{multicol}
\usepackage[bookmarks=true]{hyperref}
\usepackage{url}

\usepackage{graphicx}
\usepackage{subcaption}
\usepackage{booktabs} 
\usepackage{xcolor}

\usepackage{amsmath}
\usepackage{amssymb}
\usepackage{mathtools}
\usepackage{amsthm}

\usepackage{threeparttable}
\usepackage{tabularx}
\usepackage{adjustbox}
\usepackage{diagbox}
\usepackage{pifont}
\usepackage{wrapfig}
\usepackage{multirow}
\usepackage{makecell}
\usepackage{xspace}

\newcommand{\cmark}{\ding{51}} 
\newcommand{\xmark}{\ding{55}} 

\usepackage[capitalize,noabbrev]{cleveref}
\crefname{equation}{Eq.}{Eqs.}
\crefname{theorem}{Theorem}{Theorems}
\Crefname{theorem}{Theorem}{Theorems}
\crefname{paragraph}{paragraph}{paragraphs}
\Crefname{paragraph}{Paragraph}{Paragraphs}

\theoremstyle{plain}

\theoremstyle{definition}
\newtheorem{definition}{Definition}

\theoremstyle{remark}

\graphicspath{{figures/}}

\renewcommand{\emph}[1]{\textit{#1}}

\begin{document}

\title{\raisebox{-0.2ex}{\includegraphics[height=1.2em]{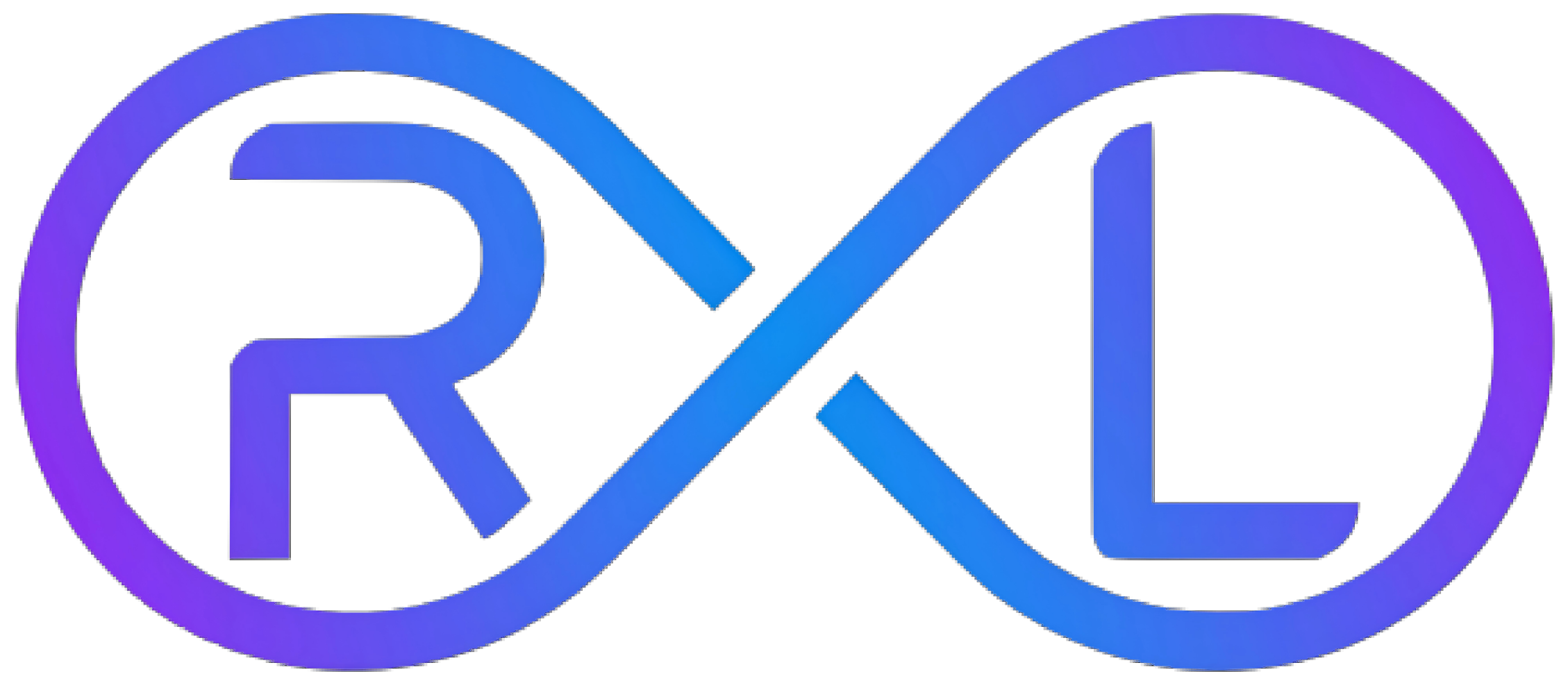}}\hspace{0.2em}RLinf-VLA: A Unified and Efficient Framework for Reinforcement Learning of Vision-Language-Action Models}
\author{
    Hongzhi Zang$^{1,*}$,
    Mingjie Wei$^{6,2,*}$,
    Si Xu$^{3}$,
    Yongji Wu$^{5}$,
    Zhen Guo$^{3}$,
    Yuanqing Wang$^{4,3}$,
    Hao Lin$^{3}$,
    Peihong Wang$^{2}$,
    \\
    Hua Yuan$^{3}$,
    Yixian Zhang$^{1}$,
    Liangzhi Shi$^{1}$,
    Yuqing Xie$^{1}$,
    Zhexuan Xu$^{1}$,
    Zhihao Liu$^{7,2}$,
    Kang Chen$^{4,2}$,
    Wenhao Tang$^{1}$,
    \\
    Quanlu Zhang$^{3}$,
    Weinan Zhang$^{6}$,
    Chao Yu$^{1,\S,\dag}$,
    Yu Wang$^{1,\dag}$
    \\[0.5em]
    {\small $^{1}$Tsinghua University \quad
    $^{2}$Zhongguancun Academy \quad
    $^{3}$Infinigence AI \quad
    $^{4}$Peking University \quad
    $^{5}$UC Berkeley}
    \\
    {\small $^{6}$Harbin Institute of Technology \quad
    $^{7}$Institute of Automation, Chinese Academy of Sciences}
    \\[0.3em]
    {\small $^*$Equal contribution \quad $^\dag$Corresponding Authors: \texttt{zoeyuchao@gmail.com}, \texttt{yu-wang@tsinghua.edu.cn} \quad $^\S$Project Lead}
    \\[0.3em]
    {\small
    \href{https://huggingface.co/RLinf}{\raisebox{-0.3ex}{\includegraphics[height=1em]{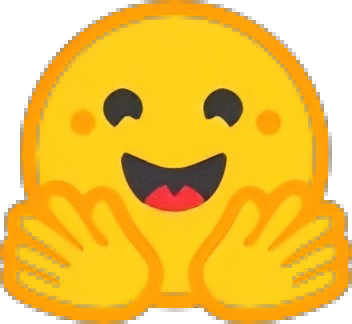}} https://huggingface.co/RLinf}
    \quad
    \href{https://github.com/RLinf/RLinf}{\raisebox{-0.3ex}{\includegraphics[height=1em]{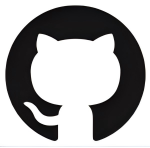}} https://github.com/RLinf/RLinf}
    }
}
\makeatletter
    \let\@oldmaketitle\@maketitle%
    \renewcommand{\@maketitle}{
    \@oldmaketitle
    \centering
    \includegraphics[width=\textwidth]{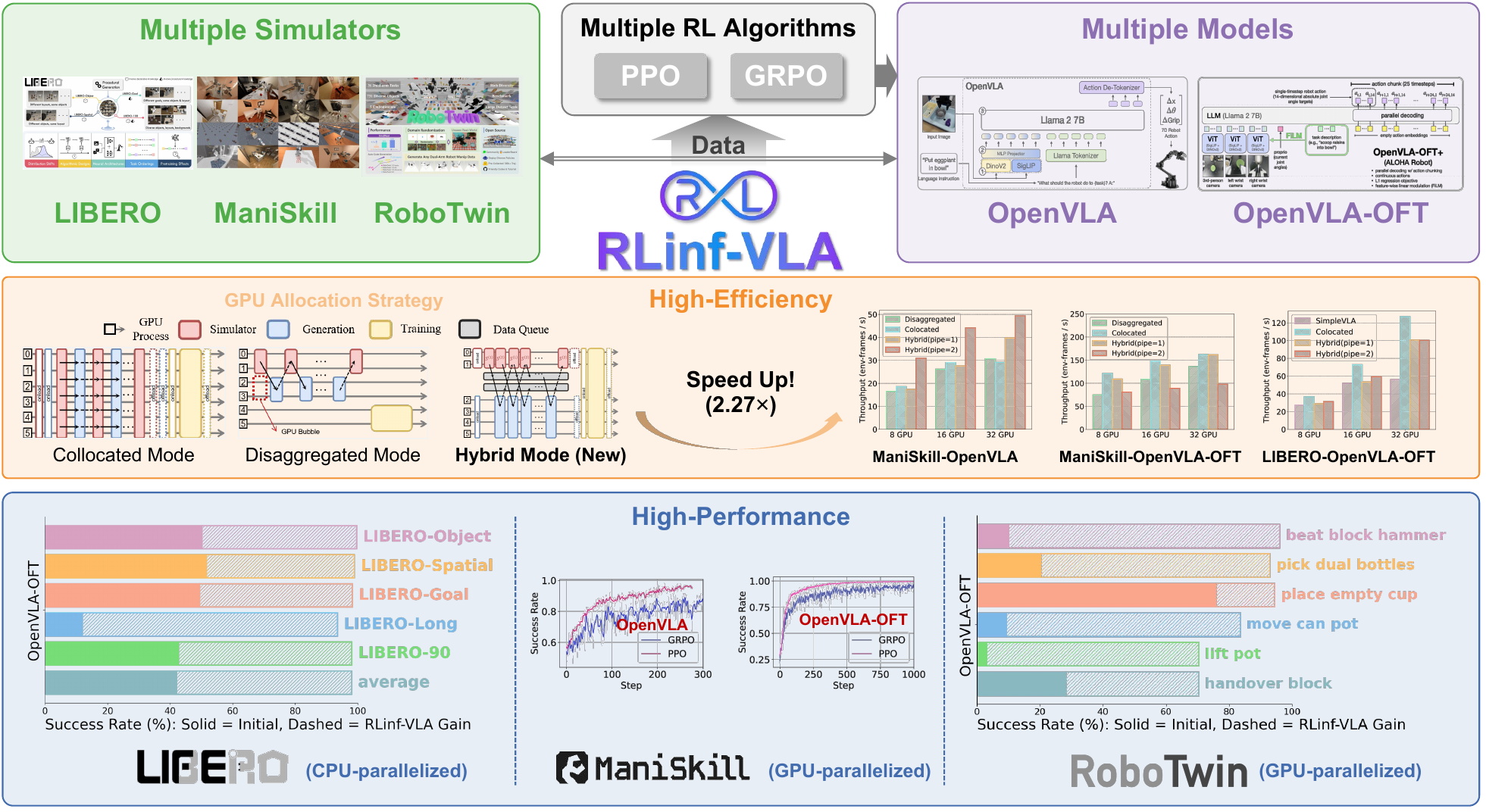}
    \captionof{figure}{
    Overview of RLinf-VLA. Built on a unified interface, RLinf-VLA seamlessly supports diverse VLA architectures, multiple RL algorithms, and various simulators. It provides three GPU allocation modes: \textit{collocated}, \textit{disaggregated}, and a novel \textit{hybrid} mode and speeds up training by $2.27\times$ compared to the baseline. A single model achieves 98.11\% success across 130 LIBERO tasks, and another single model achieves 97.66\% success across 25 ManiSkill tasks. For RoboTwin, six task-specific models achieve an average success rate of 84.63\% across 6 tasks.
    }
    \label{fig:Overview}
    }
\makeatother

\maketitle

\begin{abstract}
Recent studies have demonstrated the potential of reinforcement learning (RL) to improve the task performance of vision-language-action (VLA) models through interaction. However, current efforts remain fragmented, lacking a unified platform for fair comparison across architectures and algorithms, as well as an efficient system design for scalable training. Therefore, we present RLinf-VLA, a unified and efficient framework for scalable RL training of VLA models. RLinf-VLA standardizes the integration of diverse VLA architectures, RL algorithms, and heterogeneous simulators through a unified interface, enabling extensibility and reproducibility. To improve efficiency, the framework adopts a flexible resource allocation architecture for rendering, inference, and training in RL pipelines. In particular, RLinf-VLA introduces a hybrid fine-grained pipeline allocation strategy that achieves a 1.61$\times$–1.88$\times$ training speedup on ManiSkill. Using this framework, RL-trained models achieve strong performance across embodied benchmarks, including 98.11\% success on 130 LIBERO tasks, 97.66\% success on 25 ManiSkill tasks, and 84.63\% average success across 6 RoboTwin tasks. In addition, RLinf-VLA distills a set of effective practices for RL-based VLA training. We envision RLinf-VLA as a foundational framework for efficient, unified, and reproducible research in embodied intelligence.
\end{abstract}

\IEEEpeerreviewmaketitle

\section{Introduction}
Vision-Language-Action (VLA) models represent a new generation of foundation models that aim to unify perception, language understanding, and embodied control \citep{ma2024survey,firoozi2025foundation}. By leveraging vision-language models pretrained on internet-scale data and further training them on large, heterogeneous robot demonstration datasets \citep{open-x,khazatsky2024droid}, VLAs can map raw sensor observations and natural language instructions directly to robot actions. This paradigm has demonstrated strong generalization across a wide range of tasks \citep{team2024octo,kim24openvla,liu2024rdt,wen2025dexvla,shah2023lm,black2024pi0visionlanguageactionflowmodel,intelligence2025pi05visionlanguageactionmodelopenworld}.

However, deploying VLA models typically requires post-training to mitigate distribution mismatch between training data and deployment environments. Without effective post-training, performance can degrade significantly when models encounter novel states, instructions, or execution conditions.

Reinforcement learning (RL) has emerged as an increasingly important post-training paradigm, as it directly optimizes cumulative task rewards through trial-and-error interaction \citep{sutton1998reinforcement}. In contrast to supervised fine-tuning (SFT), another commonly used post-training approach, RL enables exploration beyond narrow expert demonstrations and equips policies with corrective and adaptive behaviors. Recent studies indicate that RL fine-tuning can yield stronger out-of-distribution generalization than SFT, particularly in terms of semantic alignment and execution robustness \citep{liu2025what}.

Despite its promise, applying RL to VLA models remains largely small-scale or fragmented \citep{liu2025what,lu2025vla}, making the development of new RL algorithms for VLAs expensive and cumbersome. Although SimpleVLA-RL~\citep{li2025simplevla} enables large-scale RL training for VLAs by building on VeRL~\citep{verl}, an RL codebase originally designed for large language models, it lacks system-level optimizations tailored to embodied settings, where simulators compete with model inference and learning for GPU resources. This limitation constrains scalability and substantially increases training time.

Moreover, online RL requires repeated and tightly coupled model–environment interactions. Without a system design explicitly tailored to this execution pattern, significant GPU idle time and pipeline bubbles can arise, leading to inefficient resource utilization. Finally, existing works often rely on disparate model architectures, RL algorithms, and evaluation protocols, making fair comparison difficult and hindering the extraction of general principles for RL-based VLA training.

To address these challenges, we present \textbf{RLinf-VLA}, a \textit{\textbf{unified}} and \textit{\textbf{efficient}} framework for scalable reinforcement learning of vision-language-action models. Our main contributions are summarized as follows:
\begin{itemize}
\item \textbf{Unified system abstraction.} RLinf-VLA provides a unified interface that supports multiple robotic simulators (e.g. ManiSkill~\citep{taomaniskill3}, LIBERO~\citep{liu2023libero}, RoboTwin~\citep{chen2025robotwin20scalabledata}), diverse VLA architectures (e.g. OpenVLA~\citep{kim24openvla}, OpenVLA-OFT~\citep{kim25openvlaoft}, $\pi_0$~\citep{black2024pi0visionlanguageactionflowmodel}), and reinforcement learning algorithms (e.g. PPO~\citep{schulman2017proximal}, GRPO~\citep{shao2024deepseekmath}). The system exposes three execution modes, including a novel hybrid GPU allocation mode, enabling scalable and configurable training across heterogeneous setups.
\item \textbf{Efficient system and algorithm design.} RLinf-VLA supports multiple flexible GPU allocation modes and further introduces a hybrid fine-grained pipelining mechanism. By improving resource allocation, RLinf-VLA substantially increases training throughput, achieving speedups of up to 2.27$\times$. In addition, the framework incorporates a set of algorithmic enhancements that further improve training efficiency and stability.
\item \textbf{Strong empirical performance and generalization.} After RL training with RLinf-VLA, a single model achieves 98.11\% success across 130 LIBERO tasks, while another single model achieves 97.66\% success across 25 ManiSkill tasks. On RoboTwin, six task-specific models achieve an average success rate of 84.63\% across 6 tasks, corresponding to an average performance improvement of 63.75\%. These results demonstrate RLinf-VLA's powerful performance in post-training.
\item \textbf{Open and extensible platform.} RLinf-VLA is released as an open-source and actively maintained platform, providing a practical foundation to accelerate, standardize, and scale reinforcement learning research for embodied intelligence.

\end{itemize}

\section{Related Work}

\subsection{Reinforcement Learning for VLA}
Vision-Language-Action models (VLAs)~\citep{kim25openvlaoft, kim24openvla, black2024pi0visionlanguageactionflowmodel, bu2025univla, intelligence2025pi05visionlanguageactionmodelopenworld, yuan2024learning, cen2025rynnvla, bi2025motusunifiedlatentaction, zheng2025x, cheang2025gr} represent a multimodal foundation model architecture designed for robotics.
RL is increasingly adopted to address the limitations of static imitation learning in VLAs. GRAPE~\citep{zhang2024grape} employs Direct Preference Optimization (DPO~\citep{rafailov2024directpreferenceoptimizationlanguage}) on partial trajectories to align models with human intent. Iterative approaches~\citep{guo2025improving, tan2025interactive} combine SFT with RL stages to balance stability and performance. VLA-RL~\citep{lu2025vla} utilizes Proximal Policy Optimization (PPO~\citep{schulman2017proximal}) to exhibit significantly stronger generalization to unseen objects and environments compared to their SFT counterparts~\citep{liu2025what}.  SimpleVLA-RL~\citep{li2025simplevla} and related variants~\citep{tan2025interactive} apply Group Relative Policy Optimization (GRPO~\citep{shao2024deepseekmath}) to improve training efficiency and performance without complex reward design. $\pi_{RL}$~\citep{chen2025pi_} implements the policy gradient algorithm to flow-matching model for the first time. $\pi^{*}_{0.6}$~\citep{intelligence2025pi} introduces RECAP to train a binarized value function and VLA.
A unified and efficient infrastructure is essential for advancing VLA via RL, yet current implementations remain fragmented~\citep{liu2025what, lu2025vla}. Although frameworks such as SimpleVLA-RL~\citep{li2025simplevla} have emerged, they largely follow generic LLM RL post-training paradigms and fail to adequately account for the system-level optimizations and interface extensibility required by embodied intelligence tasks.

\subsection{Simulators}
Simulators~\citep{li2024behavior1k, mclean2025metaworld, mees2022calvin, robocasa2024, mittal2025isaaclab, geng2025roboverse, bu2025agibot} offer scalable, safe, and controllable environments that bypass physical hardware constraints. They facilitate massive parallel data generation and automated resets, essential for training and evaluating generalizable VLA policies. Notable platforms include ManiSkill~\citep{taomaniskill3} for physics-rich manipulation, LIBERO~\citep{liu2023libero} for instruction-conditioned reasoning, and RoboTwin~\citep{chen2025robotwin20scalabledata} for bimanual tasks with domain randomization. Consequently, the construction and utilization of simulators are indispensable for RL.
However, inconsistent interfaces across simulators often necessitate extensive code restructuring. A unified interface is therefore critical to enable seamless switching between diverse simulation benchmarks.
\section{Preliminary}
\subsection{Reinforcement Learning Formulation}
To motivate the design of our training infrastructure, reinforcement learning(RL) is commonly formulated as a partially observable Markov decision process (POMDP).  A POMDP is defined by the tuple $(\mathcal{S}, \mathcal{A}, P, r, \gamma, \Omega, O)$, where $\mathcal{S}$ and $\mathcal{A}$ denote the state and action spaces, $P(s' \mid s, a)$ is the transition dynamics, $r(s, a)$ is the reward function, and $\gamma \in [0,1]$ is the discount factor. Since the agent cannot directly access the underlying state, it instead receives observations $o \in \Omega$ according to the observation function $O(o \mid s)$. In vision-based manipulation tasks, observations typically consist of multi-view images, language instructions, and proprioceptive robot states, while actions correspond to continuous robot control commands or action chunks.

At each timestep $t$, the agent samples an action $a_t \sim \pi_\theta(\cdot \mid o_t)$ based on the current observation $o_t \sim O(\cdot \mid s_t)$. The environment then executes $a_t$, transitions to the next state $s_{t+1}$ via $P$, and produces the next observation $o_{t+1}$. The goal of RL is to optimize the parameterized policy $\pi_\theta$ so as to maximize the expected cumulative discounted reward over trajectories $\tau = (s_0, a_0, \dots, s_T)$:
\begin{align}
    J(\theta) = \mathbb{E}_{\tau \sim \pi_\theta} \Bigg[ \sum_{t=0}^{T} \gamma^t r(s_t, a_t) \Bigg].
\end{align}

\subsection{Action Formulation}
Compared to RL for LLMs, a key distinction in RL for VLAs lies in the hierarchical formulation of actions. While LLMs typically operate on individual tokens~\citep{shao2024deepseekmath}, VLA policies often predict either a single action~\citep{kim24openvla} or an action \textit{chunk} containing multiple actions to ensure smooth control~\citep{zhao2023learningfinegrainedbimanualmanipulation}. This leads to a three-level hierarchy consisting of \textit{Chunk}, \textit{Atomic Action}, and \textit{Token}. Each chunk may contain multiple atomic actions, and each atomic action is composed of multiple tokens, where each token typically corresponds to a specific dimension of the robot action space (e.g., end-effector pose or joint angles).

\begin{figure}[htbp]
    \centering
    \includegraphics[width=0.5\linewidth]{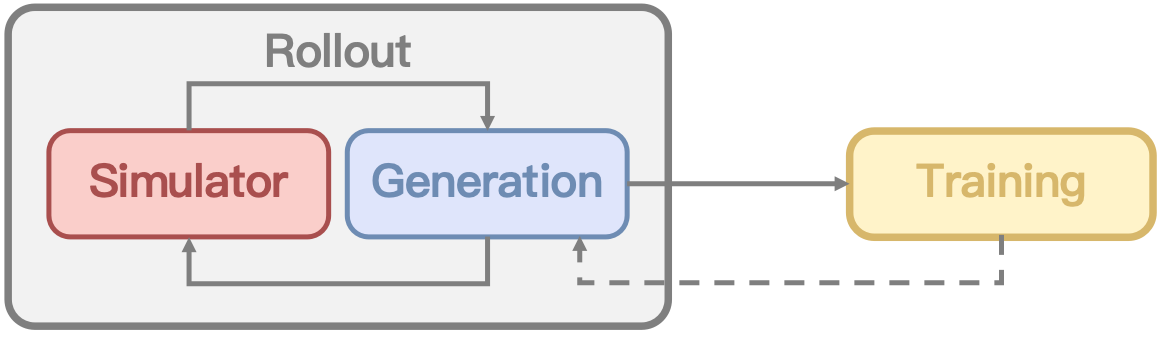}
    \caption{Pipeline of Reinforcement Learning for VLA models}
    \label{fig:rl_pipeline}
\end{figure}

\subsection{Pipeline of Reinforcement Learning for VLA models}
\label{sec:rl_pipeline}

The RL pipeline for VLA models comprises three distinct components: \textit{Generation}, \textit{Simulator}, and \textit{Training}, as illustrated in \Cref{fig:rl_pipeline}.
GPU resources are primarily allocated between the rollout phase (\textit{Generation} and \textit{Simulator}) and the optimization phase (\textit{Training}).
The interaction typically proceeds as follows~\citep{kim25openvlaoft}: the \textit{Generation} module infers an action chunk based on the current observation; the \textit{Simulator} executes this chunk and returns the observation from the final timestep. This loop continues until trajectory collection is complete, after which the \textit{Training} module updates the policy using the gathered data.

\section{Method}
\label{sec:methodology}
\Cref{sec:gpu-alloc} introduces flexible GPU Allocation Strategies for improved resource scheduling across different simulator types. \Cref{sec:unified-interface} presents a Unified Interface that seamlessly integrates diverse simulators, VLA models, and RL algorithms. Building on this system design, \Cref{sec:design-choices} distills key Design Choices for scalable RL training of VLA models.

\subsection{GPU Allocation Strategies}
\label{sec:gpu-alloc}
The RL training pipeline (described in \Cref{sec:rl_pipeline}) for VLA involves varying resource demands from \textit{Training}, \textit{Generation}, and \textit{Simulator}. Resource bottlenecks depend heavily on the simulator type: CPU-parallelized simulators are typically CPU-bound, using GPUs primarily for rendering and inference, whereas GPU-parallelized simulators execute simulation, rendering, and inference entirely on the GPU. While the latter offers higher throughput, it creates severe contention for GPU memory and compute. To maximize efficient utilization across these diverse setups, flexible and optimized GPU allocation strategies are essential. Our framework supports flexible and easily configurable allocation modes: \textit{collocated}, \textit{disaggregated}, and a novel \textit{hybrid} mode.

\subsubsection{The Collocated Mode}
\label{sec:colocated_alloc}
\begin{figure}[htbp]
    \centering
    \includegraphics[height=3.2cm]{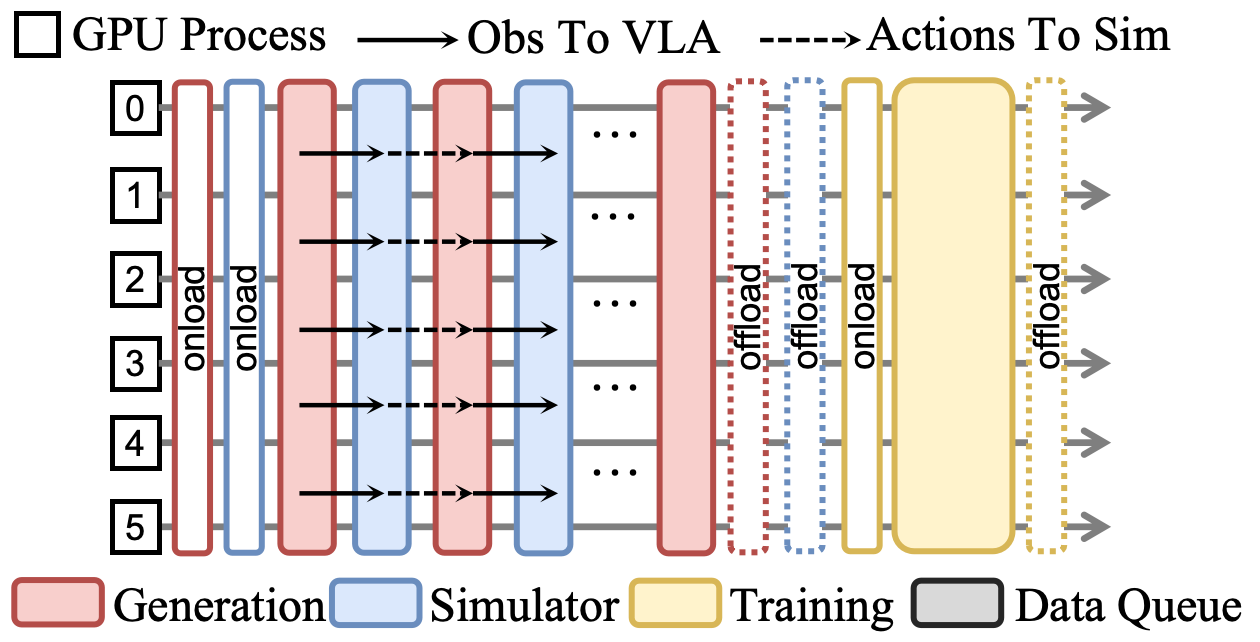}
    \caption{Collocated Mode.}
    \label{fig:gpu-alloc-collocated}
\end{figure}

In this mode (\Cref{fig:gpu-alloc-collocated}), all components co-exist on the same set of GPUs. The original version of collocated mode aimed to maximize GPU utilization by maintaining only one component on all GPUs at any given time. When a component finished its computation, it would be offloaded from GPU to CPU memory, and then onloaded again when needed for its next task. However, under the embodiment setting, both \textit{Simulator} and \textit{Generation} require GPU resources and interact frequently in an iterative manner. The overhead introduced by repeated offload-onload operations at each interaction becomes prohibitively large and unacceptable.

Therefore, we implement a modified collocated strategy where offload and onload operations for the \textit{Simulator} and \textit{Generation} occur only at the beginning and end of the rollout phase, as shown in the figure. While this approach successfully avoids the frequent offload-onload overhead during rollout, it cannot fully utilize GPU resources and remains sub-optimal. Since \textit{Generation} and \textit{Simulator} must interact iteratively, they spend significant time waiting for each other, occupying GPU memory without performing computation, leading to resource wastage and limited scalability.

\subsubsection{The Disaggregated Mode}
\begin{figure}[htbp]
    \centering
    \includegraphics[height=2.5cm]{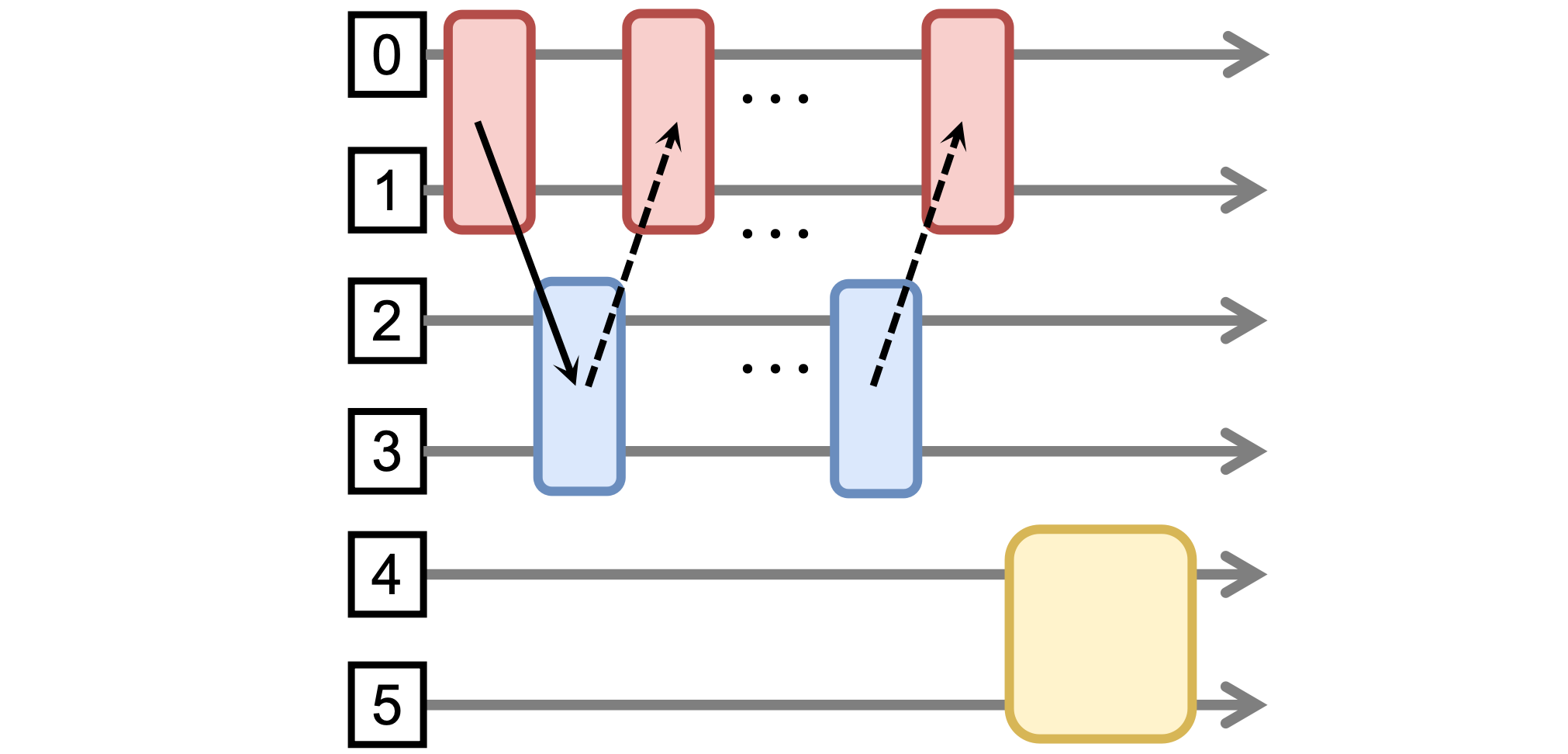}
    \caption{Disaggregated Mode (see \Cref{fig:gpu-alloc-collocated} for legend).}
    \label{fig:gpu-alloc-disagg}
\end{figure}
In this mode, each component is assigned to a distinct (possibly multi-GPU) partition, with no overlap across components. \Cref{fig:gpu-alloc-disagg} depicts the central idea. This ensures that every component can fully exploit its allocated resources. 
This mode is simple to implement, but it leads to GPU underutilization due to inter-component dependencies. For example, in \Cref{fig:gpu-alloc-disagg}, GPU 4 and 5 are completely idle during the rollout phase until training. 

\subsubsection{The Hybrid Mode with Fine-grained Pipelining}
\label{sec:arch-hybrid}
\begin{figure}[htbp]
    \centering
    \includegraphics[height=3cm]{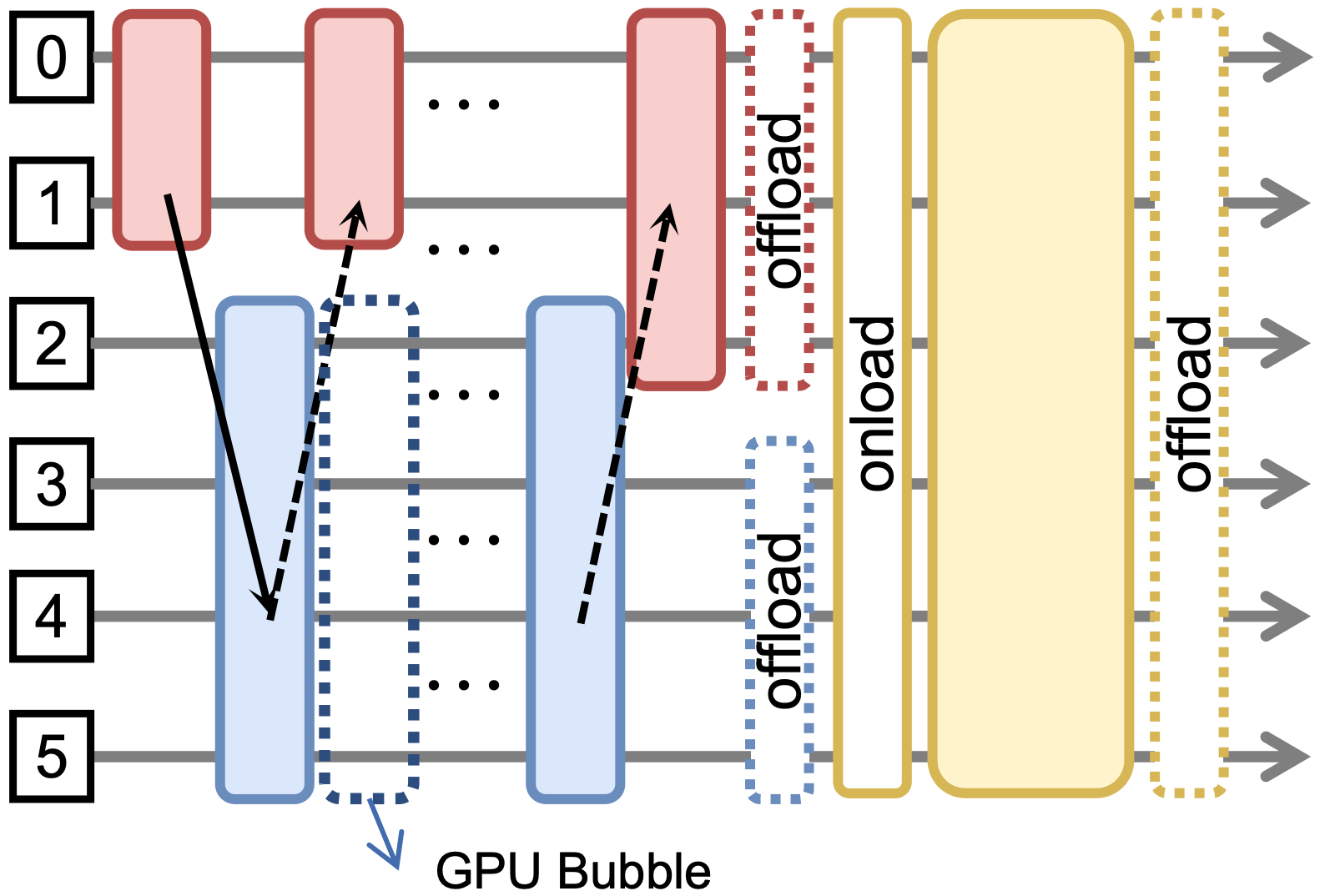}
    \caption{Hybrid Mode (see  \Cref{fig:gpu-alloc-collocated} for legend).}
    \label{fig:gpu-alloc-hybrid}
\end{figure}
To overcome the drawbacks of the above modes, we propose a hybrid mode, where each component can flexibly select GPUs. In contrast, existing distributed RL systems are less flexible. For example, veRL~\citep{sheng2024hybridflow} primarily adopts a collocated design, while AReaL~\citep{fu2025areal} and SLiME~\citep{slime_github} use a disaggregated design but still allocate GPUs to each component in contiguous blocks.

In our hybrid mode, a typical configuration is to assign \textit{Generation} and \textit{Simulator} to different GPU partitions, while allowing \textit{Training} to utilize all GPUs.  
\Cref{fig:gpu-alloc-hybrid} shows an illustration. However, the resources remain underutilized. For instance, the \textit{Simulator} must wait for actions generated by the \textit{Generation}, leaving some GPUs idle and creating ``GPU bubbles''.
\begin{figure}[htbp]
    \centering
    \includegraphics[height=3cm]{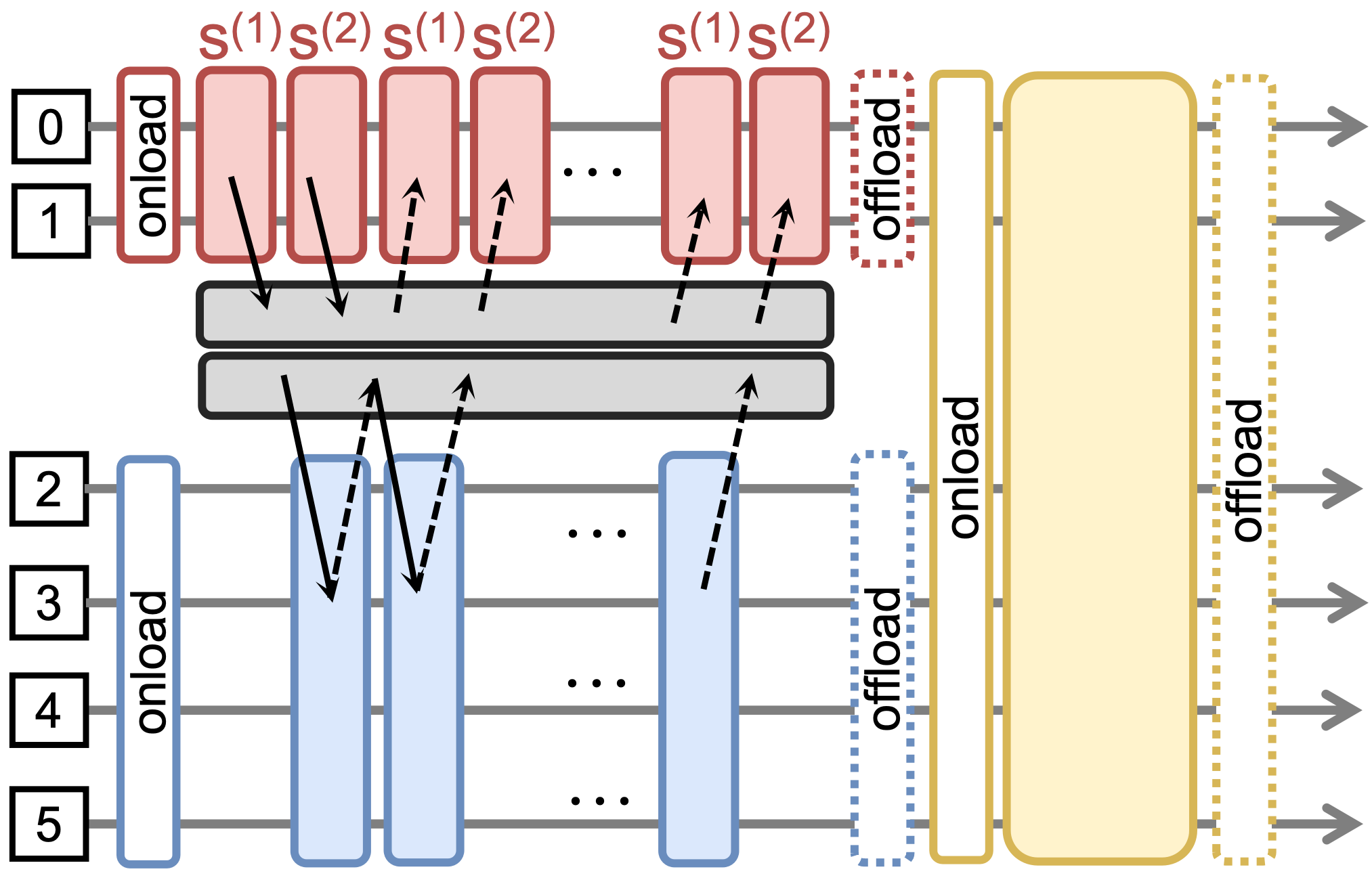}
    \caption{Hybrid Mode with Fine-grained Pipelining ($k=2$; see  \Cref{fig:gpu-alloc-collocated} for legend).}
    \label{fig:gpu-alloc-hybrid-pipe2}
\end{figure}

On top of this, we introduce fine-grained pipelining to mitigate bubbles caused by inter-component dependencies. Specifically, a simulator instance on one GPU is partitioned into multiple sub-simulators, denoted as $S^{(1)}, S^{(2)}, \ldots, S^{(k)}$. The pipeline with $k=2$ proceeds as follows, where superscripts denote simulator indices and subscripts indicate timesteps:
\begin{enumerate}
    \item At step $t=0$, $S^{(1)}$ generates the initial observation $o_0^{(1)}$, which is sent to the \textit{Generation} component to produce action $a_0^{(1)}$.  
    \item  Meanwhile, $S^{(2)}$ generates $o_0^{(2)}$ in parallel.  
    \item Once $a^{(1)}_0$ is ready, it is fed back into $S^{(1)}$ to produce the next observation $o_1^{(1)}$, while $o_0^{(2)}$ is simultaneously processed by the actor to generate $a^{(2)}_0$.  
\end{enumerate}
This scheduling allows \textit{Simulator} and \textit{Generation} to run concurrently, reducing idle time while preserving correctness.  
In this way, our framework avoids the frequent offloading overhead of collocated allocation while also eliminating the GPU bubbles that occur in disaggregated allocation.  
\Cref{fig:gpu-alloc-hybrid} provides a schematic illustration of the hybrid allocation mode with fine-grained pipelining ($k=2$).

\subsubsection{The Auto-Placement Algorithm} Since RLinf-VLA supports multiple GPU allocation modes, automatically determining an efficient allocation strategy is important. To this end, we introduce a lightweight automatic GPU placement algorithm that combines runtime profiling with allocation optimization.

We begin with several practical assumptions. Given $N$ GPUs, we assume the \textit{Training} component utilizes all $N$ GPUs whenever active, as this is typically optimal for compute-intensive workloads. We therefore focus on optimizing the \textit{Rollout} phase, which consists of the \textit{Simulator} and \textit{Generation} components. We consider GPU splitting only between these two components and restrict the pipeline depth to at most two stages. Given $N$ GPUs and $n_S$ simulation environments, our goal is to determine the GPU allocation $(N_S, N_G)$ and pipeline stage number $k \in \{1,2\}$ that minimize rollout time, where $N_S$ and $N_G$ denote the numbers of GPUs allocated to the \textit{Simulator} and \textit{Generation} components, respectively.

\noindent\textbf{Profiling.}
We estimate the execution time functions $t_G(\cdot)$ and $t_S(\cdot)$ by profiling batch sizes $\{2^0, 2^1, \ldots, n_S\}$, yielding estimates of per-chunk execution latency.

\noindent\textbf{Allocation.}
Let $m$ denote the number of chunks per epoch. We solve
$\min_{k \in \{1,2\},\; N_G + N_S = N} t_k$,
where 

Collocated ($k=1$): $t_1=
    m \left(
    t_G\!\left(\frac{n_S}{N}\right)
    +
    t_S\!\left(\frac{n_S}{N}\right)
    \right)$,

Pipelined ($k=2$):
    $t_2$ is estimated from
    $t_G(\frac{n_S}{N_G})$
    and
    $t_S(\frac{n_S}{N_S})$
    under pipelined execution.

\subsection{A Unified Interface}
\label{sec:unified-interface}
To conveniently reuse existing simulators, models, and algorithms, RLinf-VLA provides a unified interface designed for high extensibility and scalability.

\subsubsection{Multiple Simulators Support}
RLinf-VLA supports multiple robotic simulators, including ManiSkill~\citep{taomaniskill3}, LIBERO~\citep{liu2023libero}, and RoboTwin~\citep{chen2025robotwin20scalabledata}. To facilitate this, we provide a unified interface that standardizes environment interactions across different backends, such as ManiSkill and RoboTwin's GPU-parallelized environments and LIBERO's CPU-parallelized wrappers. This interface includes standard Gym-style APIs with built-in support for action chunking and flexible episode termination. Furthermore, we define a unified interface for observation and action to ensure seamless compatibility with various algorithms and models.

\subsubsection{Multiple Models Support}
Benefiting from a unified data processing pipeline across simulators, our framework facilitates the seamless integration of existing models by requiring only a standardized interaction interface. We support diverse VLA architectures, exemplified in this study by OpenVLA~\citep{kim24openvla} and OpenVLA-OFT~\citep{kim25openvlaoft}, which represent configurations with single-step and multi-step action chunking, respectively. To enable efficient fine-tuning, we integrate Low-Rank Adaptation (LoRA)~\citep{lora}.

\subsubsection{Multiple Algorithms Support}
Our framework provides support for multiple reinforcement learning algorithms, with an initial focus on Proximal Policy Optimization (PPO) \citep{schulman2017proximal} and Group Relative Policy Optimization (GRPO) \citep{shao2024deepseekmath,guo2025deepseek}. 
The core functions for on-policy RL algorithms are the advantage function and the loss function. In our framework, we can configure the algorithms by specifying these two functions only, without redundant code. 

For off-policy reinforcement learning algorithms, their application to VLA models remains an open problem, particularly in terms of achieving stable training. Nevertheless, our pipeline is fully functional and extensible to off-policy settings. As a preliminary extension, RLinf-VLA supports DSRL~\citep{wagenmaker2025steering}, a variant of Soft Actor-Critic (SAC~\citep{haarnoja2018soft}), for fine-tuning $\pi$-series models. However, off-policy methods are not the primary focus of this work, and we therefore limit our discussion.

\subsection{Design Choices for Algorithms}
\label{sec:design-choices}
RL algorithms for large models involve more complex design choices than those for smaller models. For VLAs, we adopt methodologies from both LLMs and robotics. In \Cref{sec:algo-multi-granularity}, we introduce general features applicable to all algorithms. Subsequently, in \Cref{sec:algo-detail-ppo} and \Cref{sec:algo-detail-grpo}, we detail specific design choices for PPO and GRPO, respectively. Consistent with our codebase style, all these features can be easily configured by changing values in the configuration file.



\subsubsection{Multi-Granularity Calculation Support}
\label{sec:algo-multi-granularity}

Our framework supports advantage estimation and log-probability computation at multiple levels of granularity, enabling seamless adaptation to different algorithmic requirements.

For advantage estimation, we provide two formulations for assigning advantages to action chunks $c_t = (a_{t,1}, \dots, a_{t,C})$\footnote{Here, $t$ denotes the index of the chunk, and $C$ denotes the chunk size.}. In the chunk-level formulation, the entire chunk is treated as a single macro-action and is assigned a summed reward and a shared advantage. In contrast, the action-level formulation assigns individual rewards and advantages to each atomic action $a_{t,i}$ within the chunk.

Similarly, for log-probability computation, we support three corresponding granularities:
\begin{align*}
\text{Chunk-level:} &\; \pi(c_t|o_t) = \textstyle\prod_{i=1}^C \pi(a_{t,i}|o_t, a_{t,{:}i-1}), \\
\text{Action-level:} &\; \pi(a_{t,i}|o_t, a_{t,{:}i-1}) = \textstyle\prod_{j=1}^{M} \pi(d_{t,i,j}|o_t, d_{t,i,{:}j-1}), \\
\text{Token-level:} &\; \pi(d_{t,i,j}|o_t, d_{t,i,{:}j-1}),
\end{align*}
where $d_{t,i,j}$ denotes the $j$-th token of the $i$-th atomic action.

\begin{figure}[htbp]
    \centering
    \begin{subfigure}{\linewidth}
        \centering
        \includegraphics[width=0.8\linewidth]{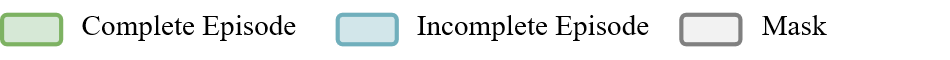}
    \end{subfigure}\\
    
    \vspace{0cm} 
    
    \begin{subfigure}{0.32\linewidth}
        \centering
        \includegraphics[width=0.95\linewidth]{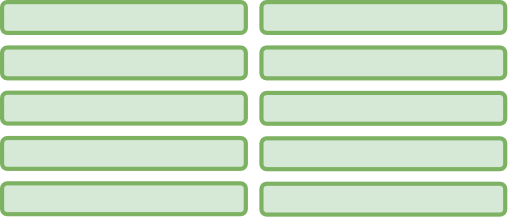}
        \caption{Fixed Length}
        \label{fig:rollout-fix}
    \end{subfigure}\hfill
    \begin{subfigure}{0.32\linewidth}
        \centering
        \includegraphics[width=0.95\linewidth]{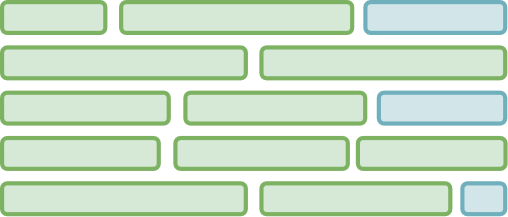}
        \caption{Partial Reset}
        \label{fig:rollout-partial}
    \end{subfigure}\hfill
    \begin{subfigure}{0.32\linewidth}
        \centering
        \includegraphics[width=0.95\linewidth]{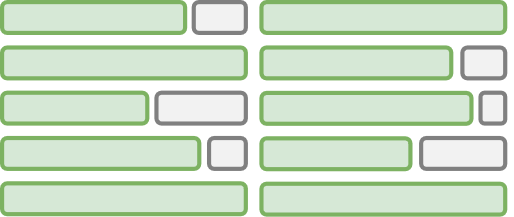}
        \caption{Valid Action Mask}
        \label{fig:rollout-mask}
    \end{subfigure}\hfill
    
    \caption{Illustration of the three rollout modes.}
    \label{fig:rollout-feature}
\end{figure}

\subsubsection{Design Choices for PPO}
\label{sec:algo-detail-ppo}
Scaling PPO to VLA models introduces several design challenges. We summarize the key design decisions adopted in our framework below.

\noindent\textbf{Partial Reset Support}
During rollouts, a sub-environment typically terminates either upon reaching the maximum episode length or upon successfully completing a task. We consider two standard strategies for handling such terminations. In the \textit{Fixed Episode Length} mode, all sub-environments are reset simultaneously only after reaching the maximum episode length. In contrast, the \textit{Partial Reset} mode resets each sub-environment immediately upon termination, independent of others. \Cref{fig:rollout-fix} and \Cref{fig:rollout-partial} illustrate the differences between these two modes.



\noindent\textbf{Critic Design}
Adapting PPO to VLA models places stringent requirements on critic design. Maintaining a separate critic network is computationally expensive and complicates GPU resource management due to the large scale of VLA models. We therefore adopt a parameter-sharing strategy between the actor and critic. Following RL4VLA~\citep{liu2025what}, we attach a lightweight value head to the language model component of the VLA for efficient state value estimation.

\noindent\textbf{Value for Action Chunks}
\label{sec:algo-detail-ppo-value}
As discussed in \Cref{sec:algo-multi-granularity}, the two advantage estimation formulations naturally induce two corresponding value estimation strategies for an action chunk $c_t = (a_{t,1}, a_{t,2}, \ldots, a_{t,C})$. In the \textit{chunk-level} formulation, the critic estimates a single scalar value for the entire chunk, treating it as a macro-action, i.e., $V: \mathcal{S} \rightarrow \mathbb{R}$. In contrast, the \textit{action-level} formulation produces a $C$-dimensional value vector, providing an individual value estimate for each atomic action $a_{t,i}$ in the chunk, i.e., $V: \mathcal{S} \rightarrow \mathbb{R}^C$. Empirically, we find that the action-level formulation consistently leads to better performance.

\subsubsection{Design Choices for GRPO}
\label{sec:algo-detail-grpo}
Transferring GRPO to embodied tasks introduces several non-trivial challenges. To address these, we adopt the following design choices.

\noindent\textbf{Valid Action Mask.}
While rollouts are executed for a fixed episode duration, tasks often terminate early. This presents two potential strategies for policy optimization: (1) using the full trajectory regardless of task completion timing, or (2) considering only timesteps prior to task completion. Our framework supports both strategies, referring to the latter as the \textit{Valid Action Mask} setting (\Cref{fig:rollout-mask}). Empirically, we find that masking out post-completion steps generally improves policy performance in the GRPO setting.

\noindent\textbf{Loss Normalization by Trajectory Length.}
\label{sec:algo-detail-grpo-norm}
To ensure that trajectories of different lengths contribute comparably to optimization, we normalize the policy loss by the number of valid timesteps in the \textit{Valid Action Mask} setting. Specifically, for a trajectory $\tau_i$ with $T_i^\text{succ}$ valid timesteps, the contribution of each timestep to the objective is scaled by $1/T_i^\text{succ}$. This prevents longer trajectories from dominating the gradient and promotes balanced learning across successful and partially completed trajectories.

\noindent\textbf{Success Rate Filter.}
\label{sec:filter}
Inspired by the dynamic sampling strategy in DAPO~\citep{yu2025dapo}, we introduce a success-rate filter for GRPO. This filter discards trajectory groups where all trajectories either succeed or fail, as computing non-zero advantages requires a mix of successful and failed outcomes. In practice, this mechanism accelerates convergence and consistently improves policy performance.

\begin{figure*}[htbp]
    \centering
    \begin{subfigure}[b]{0.23\textwidth}
    \centering
    \includegraphics[height=2.8cm]{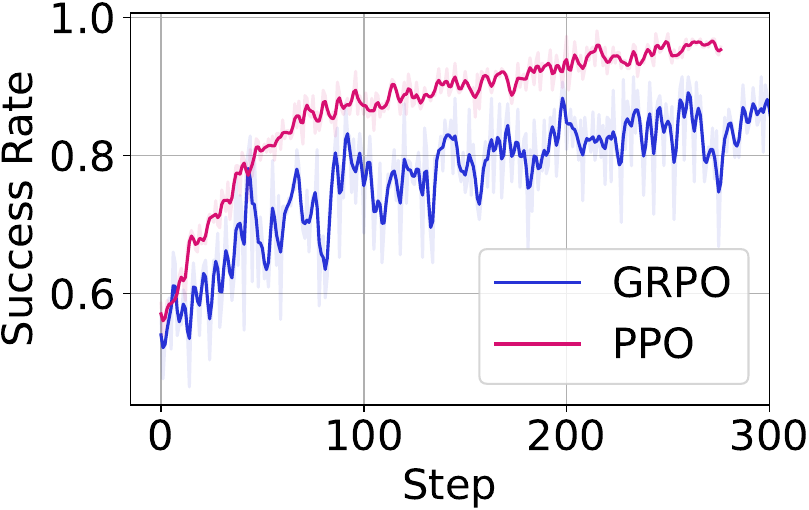}
    \caption{OpenVLA on ManiSkill.}
    \label{fig:mani-openvla}
    \end{subfigure}
    \hfill
    \begin{subfigure}[b]{0.24\textwidth}
    \centering
    \includegraphics[height=2.8cm]{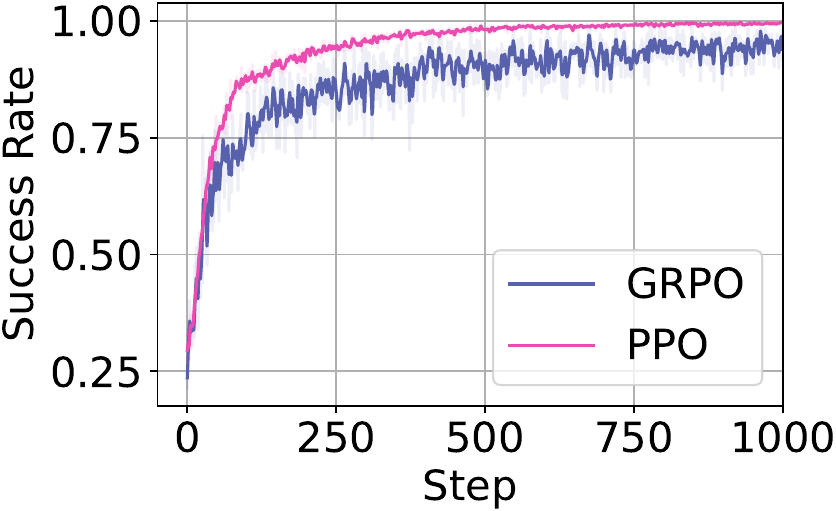}
    \caption{OpenVLA-OFT on ManiSkill.}
    \label{fig:mani-oft}
    \end{subfigure}
    \begin{subfigure}[b]{0.23\textwidth}
    \centering
    \includegraphics[height=2.8cm]{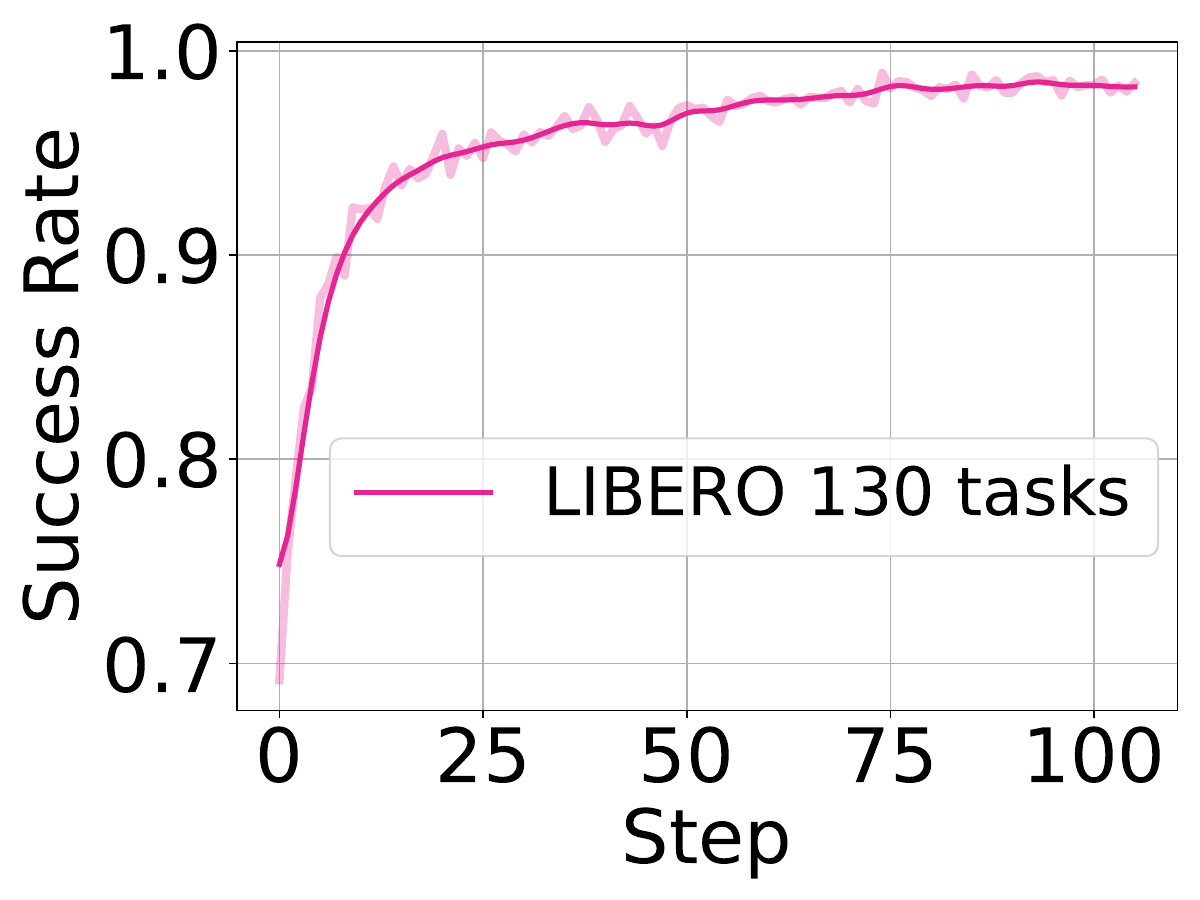}
    \caption{OpenVLA-OFT on LIBERO.}
    \label{fig:libero}
    \end{subfigure}
    \hfill
    \begin{subfigure}[b]{0.28\textwidth}
    \centering
    \includegraphics[height=2.8cm]{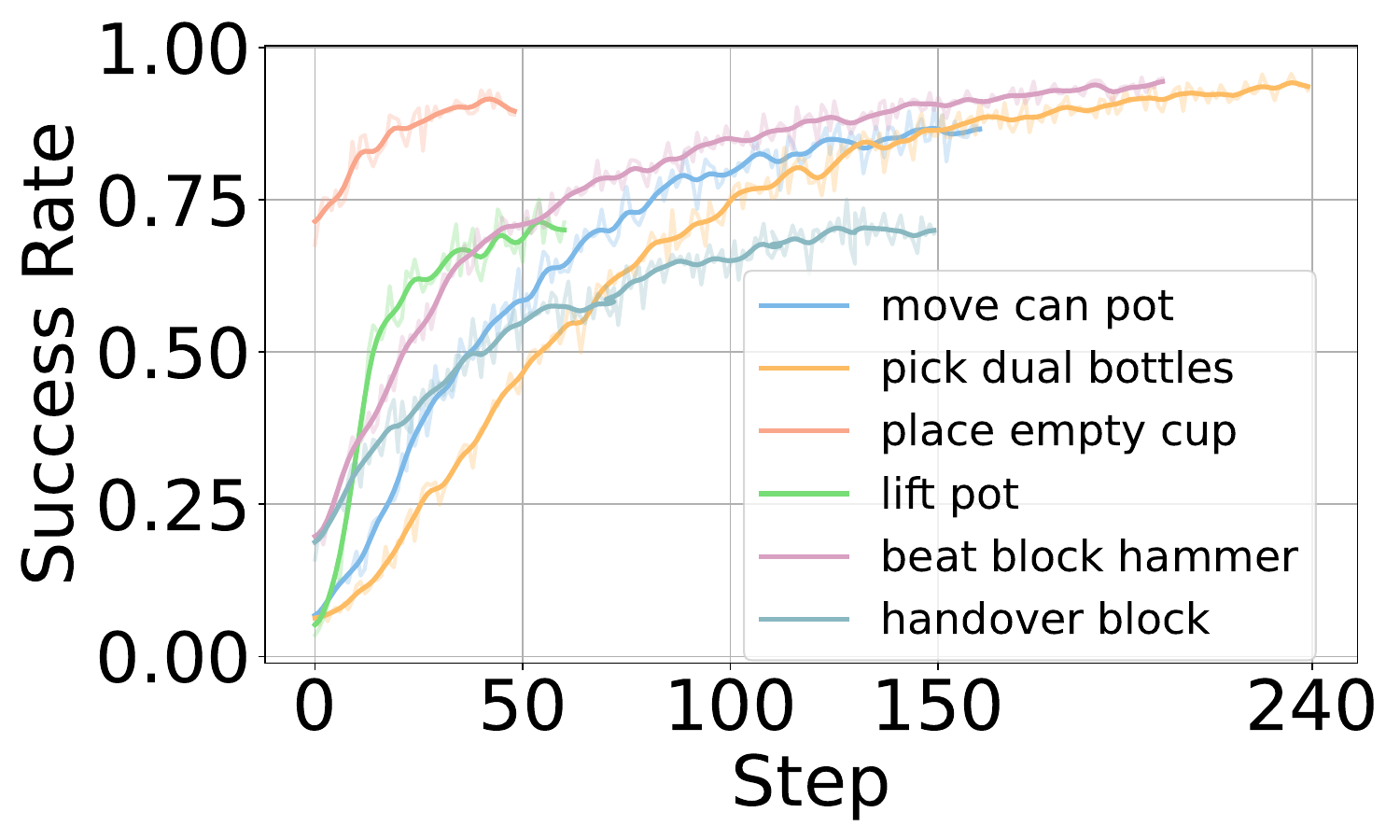}
    \caption{OpenVLA-OFT on RoboTwin.}
    \label{fig:robotwin-train}
    \end{subfigure}
    \caption{Training curves on different benchmarks. The x-axis shows the number of training epochs, and the y-axis indicates the corresponding success rate. Light-colored lines represent the raw data, while the dark-colored curves are smoothed using a Gaussian filter with $\sigma=1$. }
    \label{fig:training-curves}
    \vspace{-4mm}
\end{figure*}

\section{Experiment Results}
In this section, we systematically investigate five key questions concerning the effectiveness of the proposed RLinf-VLA framework:

\textbf{(1) Is RLinf-VLA \textit{high-performance}?}  
We evaluate RLinf-VLA on three representative testbeds, LIBERO, ManiSkill, and RoboTwin. The results demonstrate that RLinf-VLA achieves approximately 20–85\% improvement over the base model, highlighting its strong capability to support large-scale multi-task learning.

\textbf{(2) Is RLinf-VLA \textit{high-efficiency}?}  
We benchmark the framework across both GPU-parallelized and CPU-parallelized simulators, and observe that the optimal configuration improves training throughput, with speedups of up to $1.88\times$. This finding underscores the necessity of supporting diverse allocation modes. Notably, RLinf-VLA achieves up to $2.27\times$ speedup over existing frameworks.

\textbf{(3) What are the \textit{actionable practices} for applying PPO and GRPO to VLA training?}  
Through extensive ablation studies, we identify the key factors that govern training performance, offering practical guidelines for effectively deploying RL in VLA settings. 

\textbf{(4) Can simulation-based RL benefit real-world deployment?}  
We conduct a preliminary real-world deployment study to examine whether policies improved through RL in simulation can transfer to physical robots. Compared with SFT-only baselines, RL post-training leads to improved real-world performance in our setup, suggesting the potential practical value of simulation-based RL for embodied learning.

\textbf{(5) How extensible is RLinf-VLA across models and algorithms?}  
We further evaluate RLinf-VLA on additional VLA architectures and RL algorithms, including $\pi$-series models and off-policy methods. The results in \Cref{app:extensible} demonstrate the extensibility of the framework across diverse training settings.

\begin{table}[t]
\centering
\caption{Evaluation results across three simulation benchmarks. Values denote success rates (\%).}
\label{tab:eval-main-results}
    \begin{subtable}{\linewidth}
        \centering
        \caption{Evaluation on ManiSkill.}
        \label{tab:mani-eval}
        \renewcommand{\arraystretch}{1.2} 
        
        \resizebox{1.0\linewidth}{!}{
            \begin{tabular}{lcc}
            \toprule
            Method & In-Distribution   & OOD Avg. \\
            \midrule
            OpenVLA (Base)          & 53.91          & 39.10 \\
            OpenVLA (RLinf-GRPO)     & 84.38          & 75.15 \\
            OpenVLA (RLinf-PPO)      & 96.09          & \textbf{81.93} \\
            \midrule
            OpenVLA-OFT (Base)      & 28.13          & 18.29 \\
            OpenVLA-OFT (RLinf-GRPO) & 94.14          & 60.64 \\
            OpenVLA-OFT (RLinf-PPO)  & \textbf{97.66} & 77.05 \\
            \bottomrule
            \end{tabular}
        }
    \end{subtable}
    \\
    \vspace{2mm}
    \begin{subtable}{\linewidth}
        \centering
        \caption{Evaluation on LIBERO.}
        \label{tab:libero-eval}
        \renewcommand{\arraystretch}{1.} 
        \resizebox{1.0\linewidth}{!}{
            \setlength{\tabcolsep}{3pt}
            \begin{tabular}{lccccc|c}
            \toprule
            OpenVLA-OFT & Object & Spatial & Goal & Long & 90 & Avg. \\
            \midrule
            Base      & 50.20 & 51.61 & 49.40 & 11.90 & 42.67 & 42.09 \\
            RLinf-GRPO & 99.67 & 98.93 & 98.32 & 93.55 & 98.12 & 98.11 \\
            \bottomrule
            \end{tabular}
        }
    \end{subtable}
    \\
    \vspace{2mm}
    \begin{subtable}{\linewidth}
        \centering
        \caption{Evaluation on RoboTwin.}
        \label{tab:robotwin-eval}
        \renewcommand{\arraystretch}{1.}
        \resizebox{\linewidth}{!}{
            \setlength{\tabcolsep}{4.6pt}
            \begin{tabular}{lcccccc|c}
            \toprule
            OpenVLA-OFT & Cup & Hammer & Bottles & Can & Pot & Hand & Avg. \\
            \midrule
            Base      & 75.78 & 10.15 & 20.31 & 9.37 & 3.13 & 28.13 & 24.48 \\
            RLinf-GRPO &\textbf{94.53} & \textbf{96.09} & \textbf{92.96} & \textbf{83.59} & \textbf{70.31} & \textbf{70.31} & \textbf{84.63} \\
            SimpleVLA-RL & 94.2             & 87.5               & 68.3               & 61.2          & 64.1     & 57.8  &72.18    \\ 
            \bottomrule
            \end{tabular}
        }
    \end{subtable}

\end{table}
\subsection{Policy Performance}
In this section, we evaluate the effectiveness of RLinf-VLA policies across multiple benchmarks, focusing on task success rates and generalization performance.

\textbf{Experimemt Setup.}
We evaluate RLinf-VLA on three benchmarks: ManiSkill, LIBERO, and RoboTwin, with full settings in Appendix III.A.

For ManiSkill, we train on 25 pick-and-place tasks with object and receptacle variations. We follow the out-of-distribution evaluation protocol of RL4VLA~\citep{liu2025what}, measuring generalization in \textit{vision}, \textit{language}, and \textit{action}. Each sub-setting is evaluated with 256 random episodes.
For LIBERO, we use the public task groups~\citep{liu2023libero}, including LIBERO-Spatial, Object, Goal, Long, and 90, for training and evaluation. Instead of training within a single group, we train one unified model on the combined 130 tasks over 5 groups, testing large-scale multitask learning. Performance is reported across groups, averaged over 50 episodes per task with three random seeds.
For RoboTwin~\citep{chen2025robotwin20scalabledata}, we select six tasks: ``beat block hammer'', ``move can pot'', ``place empty cup'', ``pick dual bottles'', ``lift pot'', and ``handover block''. Training uses 1,000 fixed randomized scene seeds per task, and evaluation samples 128 unseen seeds to assess out-of-distribution generalization.

\textbf{Training Results.}
\indent\Cref{fig:training-curves} presents the training curves for each setup. 
\Cref{fig:mani-openvla} and \Cref{fig:mani-oft} report the training performance on ManiSkill, utilizing OpenVLA and OpenVLA-OFT models trained via PPO and GRPO. Across all settings, RL delivers substantial performance gains, improving success rates by 45\%–70\% over the baseline. Notably, PPO consistently outperforms GRPO and exhibits greater stability for both OpenVLA and OpenVLA-OFT in the ManiSkill setting.

\Cref{fig:libero} illustrates the training curve of OpenVLA-OFT with the GRPO algorithm on 130 tasks of LIBERO. Results demonstrate that the success rate improves substantially from approximately $73\%$ to $98\%$, yielding an overall performance gain of about $30\%$. These findings underscore the effectiveness of the GRPO design in RLinf-VLA for enhancing multi-task training.

\Cref{fig:robotwin-train} depicts our training curves for OpenVLA-OFT on RoboTwin. 
The training process remains stable with minimal fluctuations. Moreover, the model converges to high performance even on tasks with extremely low initial success rates (as low as 3\%).

\begin{figure*}[htbp]
    \centering
    \begin{subfigure}{0.6\textwidth}
        \centering
        \includegraphics[height=2.6cm]{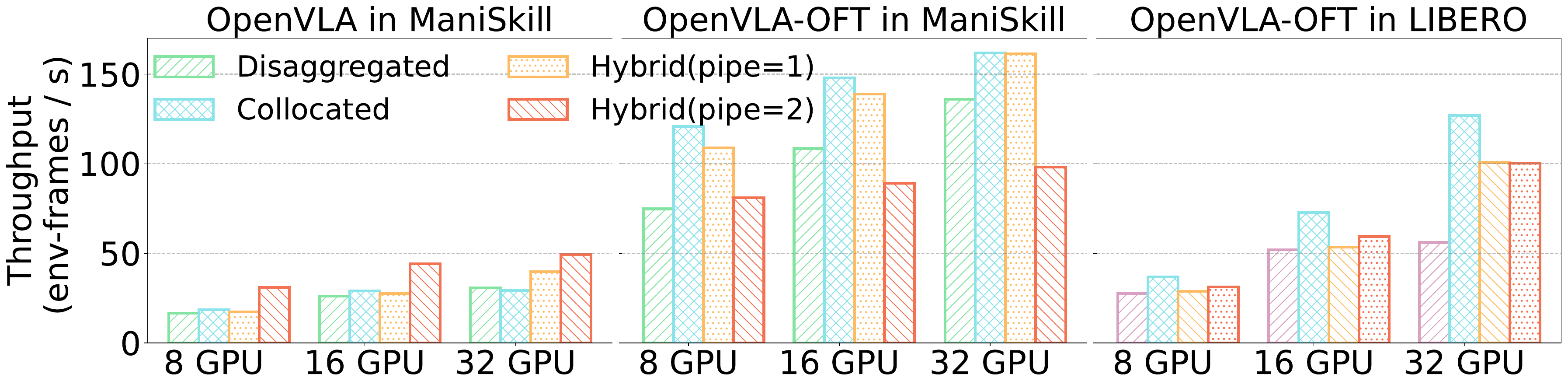}
        \caption{Throughput for different settings.}
        \label{fig:exp-efficiency}
    \end{subfigure}
    \begin{subfigure}{0.38\textwidth}
        \centering
        \includegraphics[height=2.6cm]{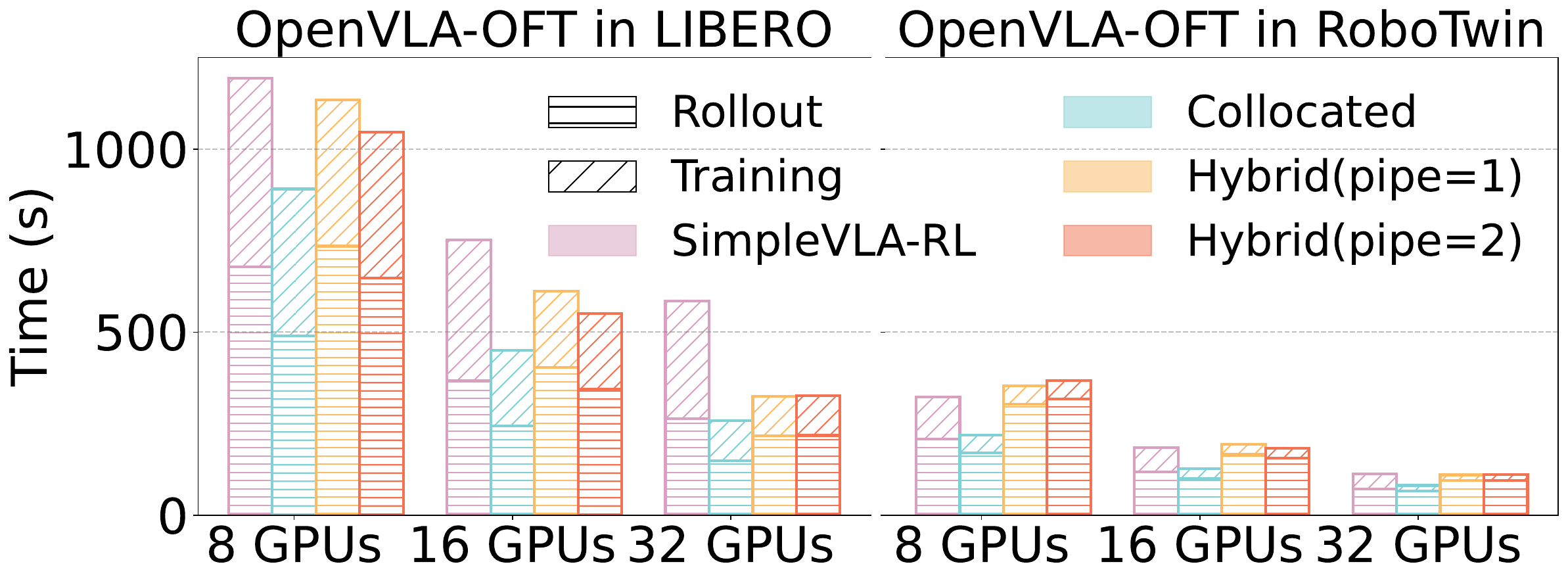}
        \caption{Latency breakdown.}
        \label{fig:latency-breakdown-both}
    \end{subfigure}
    \caption{Evaluation of system efficiency on ManiSkill, LIBERO, and RoboTwin, comparing OpenVLA and OpenVLA-OFT. Throughput (total environment frames per second) improves with increasing pipeline stages (``\texttt{pipe}'').}

    \vspace{-3mm}
    \label{fig:exp-efficiency-all}
\end{figure*}
\textbf{Evaluation Results.}
\Cref{tab:eval-main-results} summarizes RLinf-VLA results across three benchmarks, with ManiSkill detailed in \Cref{tab:mani-eval}. ``OpenVLA (Base)'' and ``OpenVLA-OFT (Base)'' denote the RL base models for OpenVLA and OpenVLA-OFT.

Direct OOD comparison between OpenVLA and OpenVLA-OFT is not strictly fair because their base performance differs. OpenVLA improves from $39.10\%$ to $81.93\%$ after RL, while OpenVLA-OFT rises from $18.29\%$ to $77.05\%$. Although starting weaker, OpenVLA-OFT shows slightly larger relative gains, indicating that base model choice strongly affects final generalization.

\Cref{tab:libero-eval} reports OpenVLA-OFT trained on 130 tasks with RLinf-VLA over five LIBERO groups. LIBERO-Object reaches $93\%$–$99\%$ success, with an overall average of $98.11\%$ and a mean improvement of $56.02\%$. Notably, we use a single unified model across all tasks, unlike standard methods that train separate models per group, demonstrating RLinf-VLA’s support for large-scale multi-task RL.

OOD results on RoboTwin are shown in \Cref{tab:robotwin-eval}. RLinf-VLA achieves an average success rate of $84.63\%$, over $60\%$ improvement. We also compare with the most relevant baseline, SimpleVLA-RL~\citep{li2025simplevla}. RLinf-VLA consistently outperforms it across all RoboTwin tasks, with particularly large gains on challenging tasks such as Bottles, Can, and Hand. These results demonstrate the effectiveness of our system design.

\subsection{System Efficiency}
In this section, we evaluate the efficiency of our system by comparing different GPU allocation modes. Our results show that the optimal strategy varies across simulators and models, highlighting the importance of flexible GPU allocation, which is a core feature of RLinf-VLA.

\textbf{Experiment Setup.}
We summarize key settings here, with full details in Appendix III.B. We evaluate representative tasks from three benchmarks. For ManiSkill, we use ``PutCarrotOnPlateInScene-v2'' with 256 parallel environments. For LIBERO, we adopt the LIBERO-Long suite, and for RoboTwin, the ``place empty cup'' task. For LIBERO and RoboTwin, parallel environments scale linearly with nodes (64, 128, and 256 for 1, 2, and 4 nodes).

We consider three GPU allocation modes (\Cref{sec:gpu-alloc}). \textit{Collocated} shares GPUs across components, \textit{Disaggregated} separates training from rollout, and \textit{Hybrid} pipelines simulation by splitting instances into stages.
For comparisons, ManiSkill lacks multi-GPU baselines, so we compare \textit{Collocated} and \textit{Hybrid} (pipelined) modes against a naive \textit{Disaggregated} baseline. For LIBERO and RoboTwin, we benchmark against SimpleVLA-RL~\citep{li2025simplevla} (collocated only) and evaluate our \textit{Collocated} and \textit{Hybrid} modes; the disaggregated mode is omitted since hybrid consistently performs better when GPUs are underutilized.

\textbf{Results.}
\Cref{fig:exp-efficiency-all} reports the end-to-end throughput of RLinf-VLA and baselines for RL training under different cluster sizes and GPU allocation modes.

From \Cref{fig:exp-efficiency}, for OpenVLA in ManiSkill, RLinf-VLA achieves up to a 1.88$\times$ speedup over the disaggregated baseline using the Hybrid (\texttt{pipe=2}) mode on 8 GPUs. Although scaling to more GPUs introduces overheads from model loading, offloading, and state switching, fine-grained pipelining reduces GPU idle time and maintains a $1.61\times$–$1.69\times$ advantage over the disaggregated mode. Deeper pipelines (\texttt{pipe=2} vs. \texttt{pipe=1}) further improve performance by reducing rollout-stage bubbles.

For OpenVLA in ManiSkill and OpenVLA-OFT in ManiSkill, as GPU count increases, the hybrid (\texttt{pipe=1}) mode outperforms the collocated mode. Since ManiSkill is GPU-parallelized, rollout throughput scales with parallel environments, making simulator resource allocation critical. To improve utilization, we enable offloading in the collocated mode (\Cref{sec:colocated_alloc}), but communication overhead grows with cluster size. In contrast, hybrid mode splits GPUs evenly between components, reducing interference.

For OpenVLA-OFT in LIBERO and ManiSkill, the behavior differs: collocated and hybrid (\texttt{pipe=1}) modes outperform disaggregated and hybrid (\texttt{pipe=2}). First, OpenVLA-OFT generates action chunks while the simulator executes them sequentially, shifting the generation-to-execution ratio from 1:1 to about 1:15. Second, LIBERO’s CPU-parallelized simulator becomes the main bottleneck. This imbalance diminishes the benefit of pipelining, effectively reverting execution to a near-sequential process.

From \Cref{fig:latency-breakdown-both}, for OpenVLA-OFT in LIBERO and RoboTwin, RLinf-VLA still achieves a 1.34$\times$–2.27$\times$ speedup over SimpleVLA-RL under collocated settings. The gains come from both rollout and training. During rollout, RLinf-VLA uses vectorized environments that are more efficient than SimpleVLA-RL’s multi-process workers. During training, system-level optimizations such as adaptive communication and the removal of redundant log-probability recomputation further reduce latency, with benefits increasing at scale.
 
\textbf{Summary.} Different simulators demand different allocation modes. RLinf-VLA’s flexible multi-mode support enables consistently high efficiency across diverse execution regimes.

\begin{figure}[htbp]
    \centering
    \includegraphics[width=0.5\linewidth]{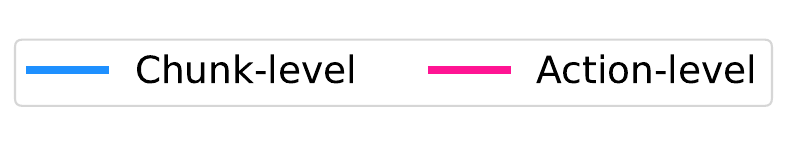}
    \\
    \vspace{-2mm}
    \begin{subfigure}{0.32\linewidth}
        \captionsetup{skip=-2pt}
        \includegraphics[width=1.15\linewidth]{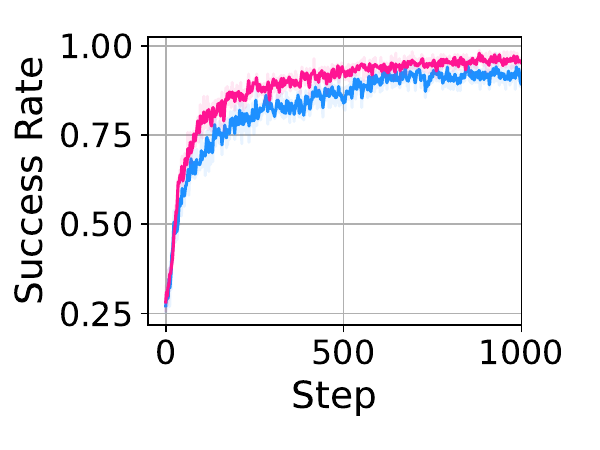}
        \caption{succ. in ManiSkill}
        \label{fig:value-mani}
    \end{subfigure}
    \begin{subfigure}{0.32\linewidth}
        \captionsetup{skip=-2pt}
        \includegraphics[width=1.15\linewidth]{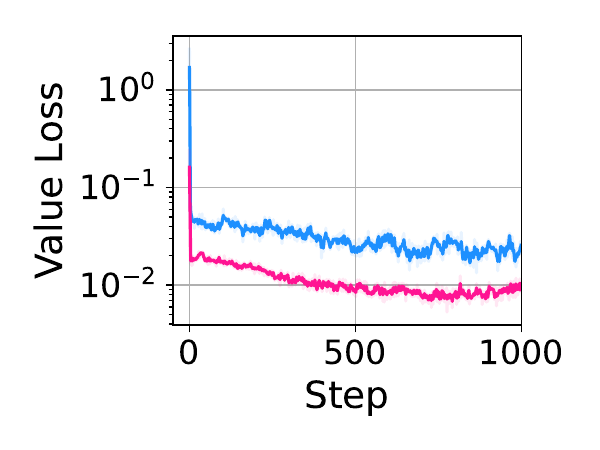}
        \caption{loss in ManiSkill}
        \label{fig:value-loss}
    \end{subfigure}
    \begin{subfigure}{0.32\linewidth}
        \captionsetup{skip=-2pt}
        \includegraphics[width=1.15\linewidth]{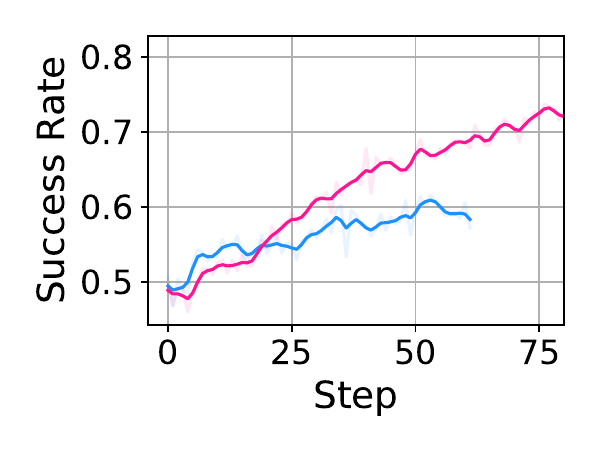}
        \caption{succ. in LIBERO}
        \label{fig:value-libero}
    \end{subfigure}
    \vspace{-1mm}
    \caption{For OpenVLA-OFT, action-level value estimation consistently outperforms chunk-level estimation across different tasks.}
    \vspace{-4mm}
    \label{fig:abl-value}
\end{figure}

\subsection{Ablation Study}
In this section, we present ablation studies under different setups and summarize key insights for PPO and GRPO training.

\subsubsection{Tips for PPO}
\label{sec:ppo-tip}

\noindent\textbf{Action-level value estimation outperforms chunk-level estimation for PPO with action chunks.}
\Cref{fig:abl-value} shows the PPO training curves comparing action-level and chunk-level value estimation as described in \Cref{sec:algo-detail-ppo-value} on the ManiSkill ``PutOnPlateInScene25Main'' task using the OpenVLA-OFT model. Action-level estimation consistently yields higher success rates and lower value loss, demonstrating more effective learning and faster policy improvement. The performance divergence between different levels of value estimation remains consistent across diverse tasks, as corroborated by similar results in the LIBERO-Goal benchmark (\Cref{fig:value-libero}).

\begin{figure}[htbp]
    \centering
    \includegraphics[width=0.6\linewidth]{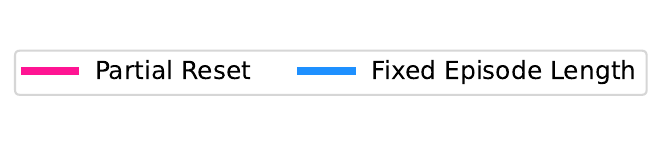}
    \\
    \vspace{-3mm}
    \begin{subfigure}{0.4\linewidth}
        \centering
        \captionsetup{skip=-2pt}
        \includegraphics[width=\linewidth]{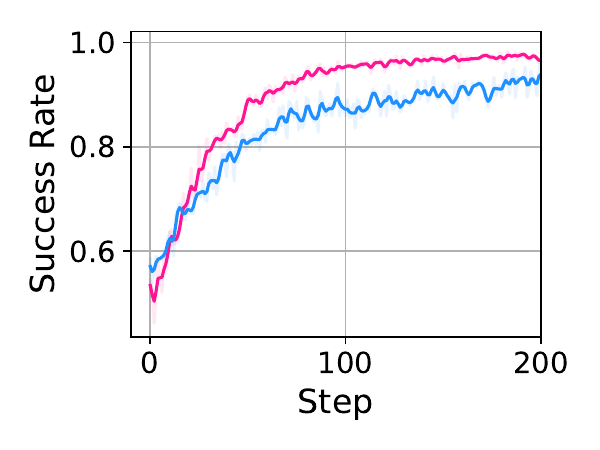}
        \caption{succ. in OpenVLA}
    \end{subfigure}
    \begin{subfigure}{0.4\linewidth}
        \centering
        \captionsetup{skip=-2pt}
        \includegraphics[width=\linewidth]{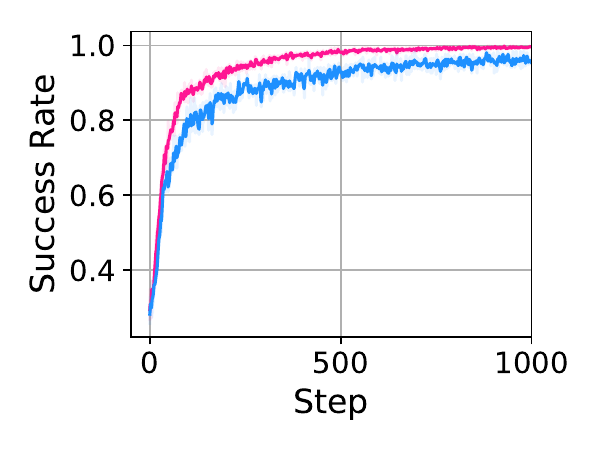}
        \caption{succ. in OpenVLA-OFT}
    \end{subfigure}
    \centering
    \vspace{-1mm}
    \caption{Partial reset consistently outperforms Fixed Episode Length mode in PPO.}
    \vspace{-4mm}
    \label{fig:abl-partial}
\end{figure}
\noindent\textbf{Partial reset substantially improves sample efficiency.} 
\Cref{fig:abl-partial} presents the PPO training curves for the \textit{Partial Reset} and \textit{Fixed Episode Length} rollout modes discussed in \Cref{sec:algo-detail-ppo}. Since the optimization objective is to achieve success at least once per rollout (``success\_once''), \textit{Partial Reset} leads to a significantly higher success rate. For a given number of training steps, the success rate under \textit{Partial Reset} consistently exceeds that of the \textit{Fixed Episode Length} mode. This trend is observed regardless of the model type (OpenVLA or OpenVLA-OFT).

\begin{figure}[htbp]
  \centering
  \vspace{-2mm}
  \includegraphics[width=0.4\linewidth]{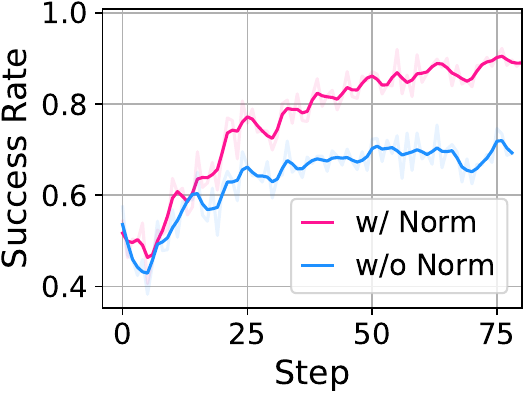}
  \caption{Ablation study on trajectory length normalization. Success rate in LIBERO-Goal with OpenVLA-OFT.}
  \vspace{-2mm}
  \label{fig:abl-norm}
\end{figure}

\subsubsection{Tips for GRPO}
\label{sec:grpo-tip}
\noindent\textbf{Trajectory length normalization. }
As discussed in \Cref{sec:algo-detail-grpo-norm}, normalizing the loss by trajectory length aims to reduce bias when episodes vary in length. \Cref{fig:abl-norm} shows that incorporating trajectory length normalization (``w/ Norm'') can lead to substantially higher performance compared with the unnormalized setting (``w/o Norm'').

\begin{figure}[htbp]
\centering
\includegraphics[width=\linewidth]{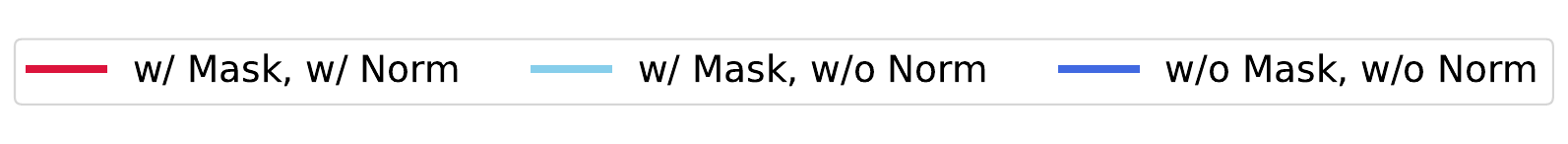}
\\
\vspace{-2mm}
\begin{subfigure}{0.45\linewidth}
\captionsetup{skip=-2pt}
\includegraphics[width=\linewidth]{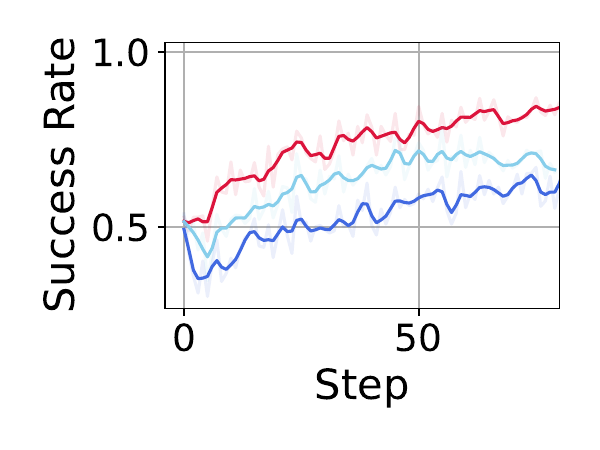}
\centering
\caption{succ. in LIBERO-Goal}
\label{fig:abl-mask}
\end{subfigure}
\begin{subfigure}{0.45\linewidth}
\captionsetup{skip=-2pt}
    \includegraphics[width=\linewidth]{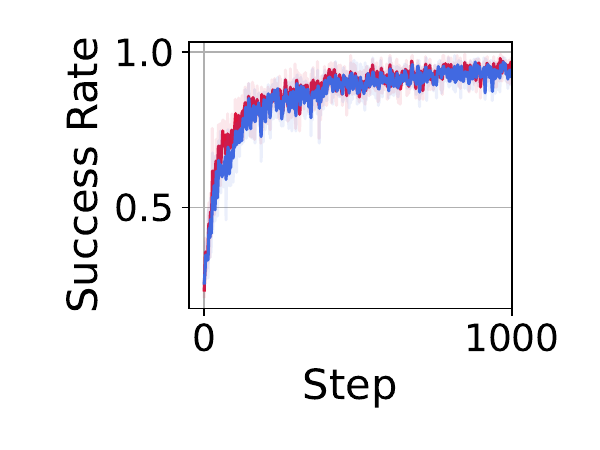}
    \centering
    \caption{succ. in ManiSkill}
\label{fig:abl-mask-mani}
\end{subfigure}
\vspace{-1mm}
\caption{Ablation studies on valid action mask in GRPO with OpenVLA-OFT.}
\vspace{-4mm}
\end{figure}

\noindent\textbf{Valid action mask. }
We also study the effect of valid action masking, introduced in \Cref{sec:algo-detail-grpo}. Since the optimization objective is defined with respect to ``success\_once'', excluding actions after reaching the success state improves sample efficiency and avoids redundant updates. Moreover, applying the valid action mask naturally results in shorter trajectories, which further benefits trajectory length normalization. \Cref{fig:abl-mask} shows results on LIBERO-Goal. Comparing the ``w/o Mask, w/o Norm'' and ``w/ Mask, w/o Norm'' curves, the setting with a valid action mask achieves consistently better performance. The ``w/ Mask, w/ Norm'' curve demonstrates that combining the mask with trajectory length normalization provides an additional improvement. However, the effect of these techniques is task-dependent. In the ManiSkill setting, we do not observe clear benefits from either valid action masking or trajectory length normalization.

\emph{(c) Success rate filtering can improve training stability in some settings of GRPO.}
As discussed in \Cref{sec:filter}, we implement a success rate filter for GRPO that discards groups in which all trajectories have identical cumulative rewards. This mechanism can improve the stability of GRPO training. For instance, in the OpenVLA ManiSkill setting, training without the filter (``w/o Filter'') exhibits a clear collapse around step 400, whereas enabling the filter (``w/ Filter'') largely alleviates this issue. However, the benefit of the filter is not universal: in the OpenVLA-OFT ManiSkill setting and the OpenVLA-OFT LIBERO-Goal setting, its effectiveness is much less pronounced.

\begin{figure}[h]
    \centering
    \includegraphics[width=\linewidth]{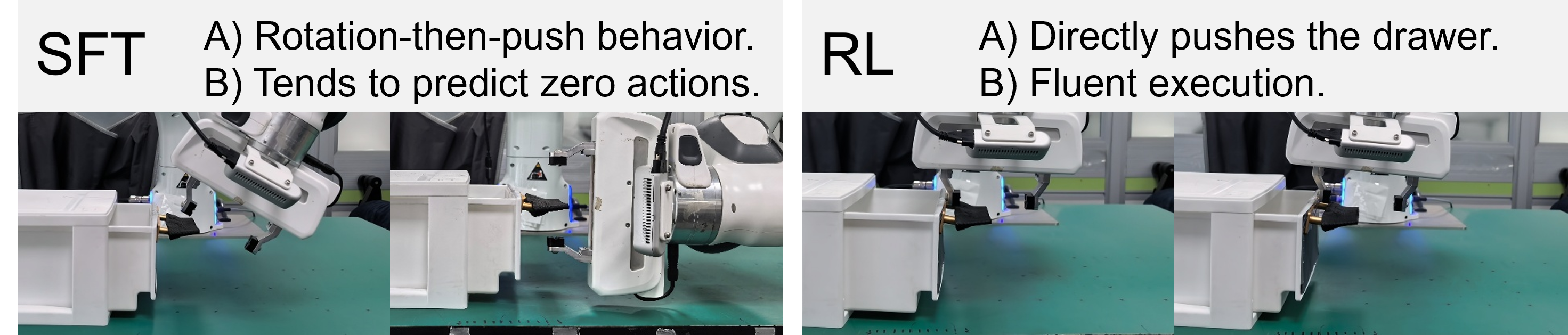}
    \caption{Deployment in the real world. }
    \label{fig:deploy}
\end{figure}
\subsection{Real-World Experiment}
We conduct a preliminary real-world experiment to evaluate whether RL improvements obtained in simulation can transfer to physical deployment.

Specifically, we consider a drawer-closing task using OpenVLA and compare an SFT-only model with a model further optimized through RL in simulation. The SFT model is trained using 20 real-world expert demonstrations together with 1000 simulated trajectories. The RL model is initialized from the SFT checkpoint and further trained with RL in simulation. To reduce the sim-to-real gap, the camera pose and robot arm configuration in the real-world setup are calibrated to match the simulation environment.

In real-world deployment, both the SFT and RL models achieve a success rate of 8/10. However, the RL model completes the task more efficiently (Fig.~\ref{fig:deploy}). We observe that the SFT model tends to imitate suboptimal human teleoperation behaviors, such as rotating before pushing the drawer, and occasionally predicts near-zero actions. In contrast, the RL-trained policy learns a more direct and fluent pushing behavior. Although limited in scale, these results suggest that simulation-based RL can improve the quality and efficiency of real-world policy execution.

\section{Conclusion}
In this work, we introduced RLinf-VLA, a unified and efficient framework for reinforcement learning-based training of Vision-Language-Action models. RLinf-VLA integrates multiple simulators, algorithms, and VLA architectures, while providing flexible execution modes and system-level optimizations that significantly improve training efficiency. Extensive experiments demonstrate that models trained with RLinf-VLA achieve high performance on a wide range of simulated tasks. Moreover, our study distills actionable practices for both PPO and GRPO, guiding future research in RL-based VLA training. By open-sourcing RLinf-VLA with ongoing maintenance, we provide the community with a foundation to accelerate, standardize, and scale research in embodied intelligence.

\section*{Acknowledgments}
This work was supported by National Natural Science Foundation of China
(No.62406159, 62325405), Zhongguancun Academy (Grant No. C20250301), Beijing National Research Center for Information Science, Technology (BNRist) and Beijing Innovation Center for Future Chips.


\bibliographystyle{plainnat}
\bibliography{paper}

\onecolumn

\begin{center}
\LARGE\textbf{Appendix for RLinf-VLA}
\end{center}
\vspace{1em}

\section{Theoretical Background}

\subsection{Definitions of RL}
\begin{definition}[Return]
The return $R_t$ is the $\gamma$-discounted cumulative reward starting from timestep $t$:
\begin{align}
    R_t = \sum_{k=0}^{\infty} \gamma^k r(s_{t+k}, a_{t+k}).
\end{align}
\end{definition}

\begin{definition}[Value Function]
The value function $V_\pi(s)$ is the expected return when starting from state $s$ and following policy $\pi$:
\begin{align}
    V_\pi(s) = \mathbb{E}_\pi[R_t \mid s_t = s],
\end{align}
where $t$ denotes an arbitrary timestep at which the agent is in state $s$.
\end{definition}

\begin{definition}[Action-Value Function]
The action-value function $Q_\pi(s, a)$ is the expected return when executing action $a$ in state $s$ and thereafter following policy $\pi$:
\begin{align}
    Q_\pi(s, a) = \mathbb{E}_\pi[R_t \mid s_t = s, a_t = a],
\end{align}
where $t$ denotes an arbitrary timestep at which the agent is in state $s$ and takes action $a$.
\end{definition}

\begin{definition}[Advantage Function]
The advantage function $A_\pi(s, a)$ quantifies how much better taking action $a$ in state $s$ is compared to the expected value of the state:
\begin{align}
    A_\pi(s, a) = Q_\pi(s, a) - V_\pi(s).
\end{align}
\end{definition}

\subsection{Introduction of PPO}
PPO~\citep{schulman2017proximal} is one of the most widely adopted reinforcement learning algorithms in robotics. PPO enhances training stability by constraining policy updates within a trust region, thereby preventing overly large changes that could destabilize learning. This is achieved through a clipped surrogate objective, which balances exploration and exploitation while maintaining sample efficiency. Due to its robustness and simplicity, PPO has become a standard baseline in robotics RL and serves as the foundation of our framework.  

In PPO, the advantage function is commonly estimated using Generalized Advantage Estimation (GAE)~\citep{Schulman2015HighDimensionalCC}:  
\begin{align}
\hat{A}_{t} = \sum_{k=0}^{T-t-1} (\gamma\lambda)^{k} \big( r_{t+k} + \gamma V(s_{t+k+1}) - V(s_{t+k}) \big),
\end{align}
where $r_{t}$ denotes the reward at timestep $t$, $V(s)$ is the value function, $\gamma$ is the discount factor, $\lambda$ controls the bias–variance trade-off, and $T$ is the episode horizon.  

The PPO optimization objective is defined as:  
\begin{equation}
J^{\text{PPO}}(\theta) = \mathbb{E}_t \Big[ 
\min \big( \rho_t(\theta) \hat{A}_t, \; \mathrm{clip}(\rho_t(\theta), 1-\epsilon, 1+\epsilon) \hat{A}_t \big) 
\Big],
\end{equation}
where
\begin{align}
\rho_{t}(\theta) = \frac{\pi_\theta(a_{t}\mid o_{t})}{\pi_{\theta_{\text{old}}}(a_{t}\mid o_{t})},
\label{equ:rho1-appendix}
\end{align}
$\pi_{\theta_{\text{old}}}$ denotes the rollout policy, and $\epsilon$ is the clipping parameter.\footnote{The definition of action $a_t$ differs across LLM and robotics RL settings. Here we treat it as a generalized action. The same applies to GRPO.}

\subsection{Introduction of GRPO}
GRPO~\citep{shao2024deepseekmath} is a recent variant of policy optimization designed to simplify reinforcement learning pipelines. Unlike PPO, which requires training both a policy model and a value model, GRPO eliminates the need for an explicit value function by leveraging group-based relative comparisons of trajectories. This design reduces the overall model complexity and avoids potential inaccuracies from value estimation. As a result, GRPO provides a lightweight yet effective alternative to PPO, making it especially attractive for large-scale VLA training where efficiency and simplicity are important. In GRPO, the important ratio can be defined as:
\begin{align}
\rho_{t}^{(i)}(\theta) = \frac{\pi_\theta(a_{t}^{(i)}\mid o_{t}^{(i)})}{\pi_{\theta_{\text{old}}}(a_{t}^{(i)}\mid o_{t}^{(i)})},
\label{equ:rho2-appendix}
\end{align}
where $i$ denotes $i$-th trajectory.
But the advantage can be defined as:
\begin{align}
\hat{A}^{(i)}=\hat{A}_{t}^{(i)} = \frac{\mathcal{R}^{(i)} - \mathrm{mean}(\{\mathcal{R}^{(j)}\}_{j=1}^G)}{\mathrm{std}(\{\mathcal{R}^{(j)}\}_{j=1}^G)},\quad \forall t\in \{0, 1, \ldots, |\tau^{(i)}|\},
\end{align}
where $\mathcal{R}^{(i)}$ denotes the total reward of trajectory $\tau^{(i)}$. Note that $\mathcal{R}$ differs from the discounted return $R$, and $G$ represents the group size. Given an initial observation $o_0$ from sample buffer $D$, the behavior model $\pi_{\theta_{\text{old}}}$ generates $G$ trajectories $\{\tau_i\}_{i=1}^{G}$.
The GRPO optimization objective is:

\begin{equation}
J^{\text{GRPO}}(\theta)
= \mathbb{E}_{o_0 \sim D,\ \{\tau^{(i)}\} \sim \pi_{\theta_{\text{old}}}}
\Bigg[\frac{1}{G}\sum_{i=1}^G\frac{1}{|\tau^{(i)}|}
\sum_{t=1}^{|\tau^{(i)}|} 
\min\Bigl(
\rho_{t}^{(i)}(\theta)\,\hat{A}^{(i)},\;
\mathrm{clip}\bigl(\rho_{t}^{(i)}(\theta),\,1-\epsilon,\,1+\epsilon\bigr)\,\hat{A}^{(i)}
\Bigr)\Bigg],
\end{equation}

where $|\tau^{(i)}|$ is the length of trajectory $\tau^{(i)}$, $\epsilon$ is the clip parameter.

\section{Implementation Details of RLinf-VLA}

\subsection{The Unified Interface}
The unified interface consists of two categories: \textit{core functions} and \textit{utility functions}.  

\paragraph{Core Functions}
We implement standard Gym-style APIs, including ``\texttt{reset}'' and ``\texttt{step}''. The ``\texttt{step}'' function additionally supports an ``\texttt{auto\_reset}'' option: when enabled, any sub-environment that terminates or is truncated will be automatically reset, thereby improving sample efficiency by avoiding idle environments. Following the ManiSkill~\citep{taomaniskill3} convention, we also support the ``\texttt{ignore\_terminations}'' option. When enabled, termination signals are ignored and only truncation signals are respected, meaning that an episode ends only when the maximum episode length is reached. This feature allows us to flexibly support different implementation variants, such as ``Partial Reset'' and ``Valid Action Mask''.

Beyond these, we extend the interface with a ``\texttt{chunk\_step}'' function to handle action chunks. Instead of naively looping over the chunk with repeated ``\texttt{step}'' calls, this function manages episode termination more carefully. Two modes are supported: (1) reset immediately when a sub-environment finishes within the chunk, or (2) defer reset until the entire chunk has been executed. This flexibility ensures the correct handling of episode boundaries when chunked actions are used.  

\paragraph{Utility Functions}
Utility functions offer convenient support for training and evaluation. For instance, visualization utilities enable effortless generation of videos during rollouts or evaluation. In addition, we provide specialized utilities required by specific algorithms. For example, GRPO necessitates that all environments within a group share the same initial state, which can be ensured by setting ``\texttt{use\_fixed\_reset\_state\_ids=True}''.

\subsection{GPU Allocation Modes}

\paragraph{Configuring GPU Allocation Modes}
Our framework exposes a simple configuration interface that allows users to flexibly choose GPU allocation strategies without explicitly specifying the mode. Instead, users only need to assign GPU IDs for each component via the following fields:
\begin{itemize}
    \item \texttt{cluster.component\_placement.env} for the \textit{Simulator},
    \item \texttt{cluster.component\_placement.rollout} for \textit{Generation},
    \item \texttt{cluster.component\_placement.actor} for \textit{Train}.
\end{itemize}

In addition, offloading can be enabled or disabled independently for each component using:
\begin{itemize}
    \item \texttt{env.enable\_offload},
    \item \texttt{rollout.enable\_offload},
    \item \texttt{actor.enable\_offload}.
\end{itemize}

Finally, fine-grained pipelining is controlled by the parameter \texttt{rollout.pipeline\_stage\_num}. A value greater than 1 splits each simulator instance on a GPU into multiple pipeline stages, while a value of 1 disables pipelining.

\subsection{Details for Design Choices}
\subsubsection{Details for Multi-Granularity Support}

\begin{figure}[htbp]
    \centering
    \includegraphics[width=0.5\linewidth]{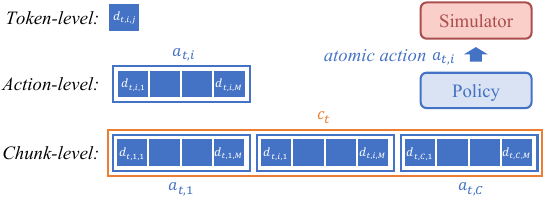}
    \caption{Illustration of different log-probability granularities.}
    \label{fig:level}
\end{figure}

In the main text, we refer to three levels of granularity: \emph{token-level}, \emph{action-level}, and \emph{chunk-level}. \Cref{fig:level} provides an illustrative example of these levels. 
An \emph{action chunk} consists of multiple atomic actions, which are the actual control signals executed by the simulator at each step. 
Each atomic action, in turn, is composed of multiple tokens, where each token corresponds to a single action dimension.

\label{sec:algo-adv-appendix}
The complete set of supported combinations between advantage computation granularity and log-probability granularity is summarized in \Cref{tab:adv-logprob-gran}.

\begin{table}[htbp]
\centering
\caption{Supported combinations of advantage and log-probability granularities.}
\label{tab:adv-logprob-gran}
\begin{tabular}{l|ccc}
\toprule
\diagbox{Advantage}{Log-Probability} & Chunk-level & Action-level & Token-level \\
\midrule
Chunk-level   & \cmark & \cmark & \cmark \\
Action-level  & \xmark & \cmark & \cmark \\
\bottomrule
\end{tabular}
\end{table}

\subsubsection{Details for Loss Normalization by Trajectory Length}
\label{sec:algo-detail-grpo-norm-appendix}
As mentioned in the main text, we normalize the policy loss by the trajectory length under the \textit{Valid Action Mask} setting. Specifically, if a trajectory $\tau_i$ has $T_i^\text{succ}$ valid timesteps, the contribution of each timestep to the objective is scaled by $1/T_i^\text{succ}$. The GRPO objective under this setting is:
\begin{equation}
J^{\text{GRPO}}(\theta) =
\mathbb{E}_{o_0 \sim D, \{\tau_i\} \sim \pi_{\theta_\text{old}}} \Bigg[
\frac{1}{G} \sum_{i=1}^G \frac{1}{T_i^\text{succ}} \sum_{t=1}^{T_i^\text{succ}} 
\min \Big( \rho_{i,t}(\theta) \hat{A}_i, \; 
\mathrm{clip}(\rho_{i,t}(\theta), 1-\epsilon, 1+\epsilon) \hat{A}_i \Big)
\Bigg].
\end{equation} 
This prevents longer trajectories from disproportionately dominating the gradient and ensures balanced learning across trajectories of varying lengths.

\section{Experiment Setup}
\begin{table}[htbp]
\centering
\caption{Hyperparameter settings for training.}
\scriptsize
\begin{threeparttable}
\resizebox{\linewidth}{!}{
\begin{tabular}{lcccccc}
\toprule
\multirow{3}{*}{Parameter} & \multicolumn{2}{c}{OpenVLA~\citep{kim24openvla}} & \multicolumn{2}{c}{OpenVLA-OFT~\citep{kim25openvlaoft}} & OpenVLA-OFT & OpenVLA-OFT \\
 & \multicolumn{2}{c}{ManiSkill~\citep{taomaniskill3}}  & \multicolumn{2}{c}{ManiSkill}  & LIBERO~\citep{liu2023libero} & RoboTwin~\citep{chen2025robotwin20scalabledata} \\
\cmidrule(lr){2-3} \cmidrule(lr){4-5} \cmidrule(lr){6-7}
 & PPO & GRPO & PPO & GRPO & GRPO & GRPO \\
\midrule
\# Parallel Envs      & 128        & 256        & 128        & 256        & 64          & 128        \\
Max Episode Steps     & 80         & 80         & 80         & 80         & 512  & 200\tnote{a}  \\
Max Steps per Rollout Epoch & 160        & 80        & 160        & 80        & 512        & 200\tnote{a}        \\
Group Size            & 1          & 8          & 1          & 8          & 8           & 8          \\
Rollout Epoch         & 1          & 1          & 1          & 1          & 64          & 8          \\
Partial Reset         & True       & False      & True       & False      & False       & False      \\
Valid Action Mask     & False      & True       & False      & True       & True        & True       \\
Action Chunk Size     & 1          & 1          & 8          & 8          & 8           & 25        \\
\midrule
Global Batch Size     & 640        & 640        & 640        & 640        & 16384       & 1024       \\
Micro Batch Size      & 40         & 40         & 40         & 40         & 32          & 32         \\
is\_lora~\citep{lora}              & True       & True       & True       & True       & ---         & True       \\
\midrule
Learning Rate         & 1e-4       & 1e-5       & 1e-4       & 1e-4       & 2e-5        & 1e-4       \\
adam\_eps             & 1e-8       & 1e-8       & 1e-8       & 1e-8       & 1e-8        & 1e-5       \\
Clip Ratio ($\epsilon$) & (0.2, 0.28) & (0.2, 0.28) & (0.2, 0.28) & (0.2, 0.28) & (0.2, 0.28)  & (0.2, 0.28) \\
$\gamma$              & 0.99       & ---        & 0.99       & ---        & ---         & ---        \\
GAE $\lambda$         & 0.95       & ---        & 0.95       & ---        & ---         & ---       \\
Temperature (train)   & 1.0        & 1.0        & 1.0        & 1.0        & 1.6         & 1.6        \\
\bottomrule
\end{tabular}
}
\begin{tablenotes}
\scriptsize
\item[a] Note that the ``pick dual bottles'' and ``handover block'' task in RoboTwin requires 100 and 400 steps respectively.
\end{tablenotes}
\end{threeparttable}
\label{tab:hyperparameter-settings-training}
\end{table}

\subsection{Details for Performance Experiments}
This part is the details of training and evaluation for High-Performance experiments, including metric for evaluation, base models selection and hyperparameter settings for training. Unless otherwise specified, all curves are smoothed using a Gaussian filter ($\sigma=1$), and success rate is computed under the ``\texttt{success\_once}'' criterion, where an episode is considered successful if the success state is reached at least once. 

\subsubsection{Base Models for Training}
For the OpenVLA (Base) in ManiSkill, we adopt the pre-trained checkpoint OpenVLA (Base) from RL4VLA~\citep{liu2025what}. 
For the OpenVLA-OFT (Base) in ManiSkill, we perform our own LoRA fine-tuning to get OpenVLA-OFT (Base) using motion planning data collected from the ``PutOnPlateInScene25Main-v3'' task. 

For OpenVLA-OFT (Base) in LIBERO 130 tasks, we fine-tune OpenVLA-OFT with LoRA using demonstrations from all five task suites in LIBERO.

For OpenVLA-OFT (Base) in RoboTwin, we adopt the pre-trained checkpoint OpenVLA-OFT (Base) from SimpleVLA-RL~\citep{li2025simplevla}. Besides, we also reuse the train/eval seeds from SimpleVLA-RL~\citep{li2025simplevla}.

\subsubsection{Hyperparameter Settings for Training}
In \Cref{tab:hyperparameter-settings-training}, we list the specific hyperparameters used for different experiments.

\subsubsection{Metric for Evaluation}
For supervised fine-tuned models, we set \texttt{do\_sample=False} during evaluation. And for RL-trained models, we set \texttt{do\_sample=True} during evaluation, and conduct three times evaluation for each model. And we report the mean and standard deviation of the success rates. (Now only the LIBERO setting has standard deviation.)

\subsection{Details for Efficiency Experiments}
\begin{table}[htbp]
    \centering
    \caption{The setting of \texttt{enable\_offload} among different GPU allocation modes.}
    \label{tab:offload-cfg}
    \resizebox{0.4\linewidth}{!}{
        \begin{tabular}[t]{l|lll}
        \toprule
                      & Simulator & Generation & Training \\
        \midrule
        Disaggregated & False     & False      & False    \\
        Colocated     & True      & True       & True     \\
        Hybrid        & True      & True       & True     \\
        \bottomrule
        \end{tabular}
    }

\end{table}

\begin{table}[htbp]
    \centering
    \caption{Hyperparameters of GPU allocation modes.}
    \label{tab:gpu-alloc-cfg}
    \resizebox{0.5\linewidth}{!}{
        \begin{tabular}{llccc}
        \toprule
        \# GPU     &  Allocation Mode   & Simulator & Generation & Training     \\ 
        \midrule
                   & Disaggregated             & 0-1       & 2-3        & 4-7   \\
        8 GPUs     & Colocated                 & 0-7       & 0-7        & 0-7   \\
                   & Hybrid                    & 0-3       & 4-7        & 0-7   \\ 
        \midrule
                   & Disaggregated             & 0-3       & 4-7        & 8-15  \\
        16 GPUs    & Colocated                 & 0-15      & 0-15       & 0-15  \\
                   & Hybrid                    & 0-7       & 8-15       & 0-15  \\ 
        \midrule
                   & Disaggregated             & 0-7       & 8-15       & 16-31 \\
        32 GPUs    & Colocated                 & 0-31      & 0-31       & 0-31  \\
                   & Hybrid                    & 0-15      & 16-31      & 0-31  \\ 
        \bottomrule
        \end{tabular}
    }
\end{table}
\subsubsection{Hyperparameters for Efficiency Experiments}
See \Cref{tab:offload-cfg} for how we enable the offload configurations and \Cref{tab:gpu-alloc-cfg} for how we configure the GPU allocation.

\subsubsection{Metric for Efficiency}
We use \textit{throughput} as the evaluation metric for efficiency. Specifically, throughput is defined as the total number of rollout environment frames divided by the total wall-clock time of one training epoch, which approximately equals the sum of rollout time and training time. 

\subsubsection{Tasks for Efficiency}
We evaluate our framework on all tasks in ManiSkill, LIBERO and RoboTwin.

For \textbf{ManiSkill}, we select the ``PutCarrotOnPlateInScene-v2'' task for a quick test. It is adapted from ``PutCarrotOnPlateInScene-v1'' in ManiSkill, with modifications to the reset function. 
In the original reset implementation, additional simulation steps were applied to ensure that all objects remained static. However, this significantly increased the reset time compared to the step function and was incompatible with partial reset. In practice, we observed that these extra steps had a negligible impact on object states in the ``PutCarrotOnPlateInScene'' task, since the initial position of the carrot was already carefully calibrated. We therefore delete the additional simulation steps in the reset function.
The evaluation is conducted on 8, 16, and 32 NVIDIA H100 (80GB) GPUs, and we use 256 parallel environments, each running for 80 steps.  

For \textbf{LIBERO}, we adopt the LIBERO-Long task set for a quick test, which is the longest task set in LIBERO. The number of parallel environments is set to 64, 128, and 256 for 8-, 16-, and 32-GPU setups, respectively, with the corresponding number of environment steps set to 4096, 2048, and 1024.\footnote{Since vectorized environments in LIBERO rely on multi-processing, the number of physical CPU cores per node becomes the upper bound for efficient rollout. Therefore, we scale the number of parallel environments in proportion to the number of nodes, and consequently to the total number of GPUs.}

For \textbf{RoboTwin}, we select the \texttt{place\_empty\_cup} task for a quick test, which requires the robot to use an arm to place the empty cup on the coaster. The number of parallel environments is set to 64, 128, and 256 for 8-, 16-, and 32-GPU setups, respectively, with the corresponding number of environment steps set to 800, 400, and 200. We have the same GPU allocation setting as LIBERO-Long, due to the GPU resources being the upper bound for the rollout.

\subsubsection{Baselines for Efficiency Experiments}
For ManiSkill, no existing framework supports multi-GPU rollout and training. Thus, we take the naive disaggregated allocation mode as the baseline, and compare it with our colocated mode and hybrid modes, where the hybrid mode includes both one-stage and two-stage fine-grained pipelining configurations.

For LIBERO and RoboTwin, we compare against SimpleVLA-RL~\citep{li2025simplevla}, an open-source framework for RL of VLA Models training built on VeRL. Since SimpleVLA-RL only supports the colocated mode, we use it as the baseline and additionally evaluate our colocated mode and hybrid mode (with one-stage and two-stage fine-grained pipelining). We omit the disaggregated mode results for LIBERO and RoboTwin, as during training the \textit{Training} component never reuses the same GPUs as the rollout process. Consequently, even when accounting for the offload–onload overhead, the hybrid (one-stage) configuration provides superior performance to the disaggregated mode.

\section{Additional Experimental Results}
\begin{figure}[htbp]
    \centering
    \begin{subfigure}{0.3\linewidth}
    \includegraphics[width=\linewidth]{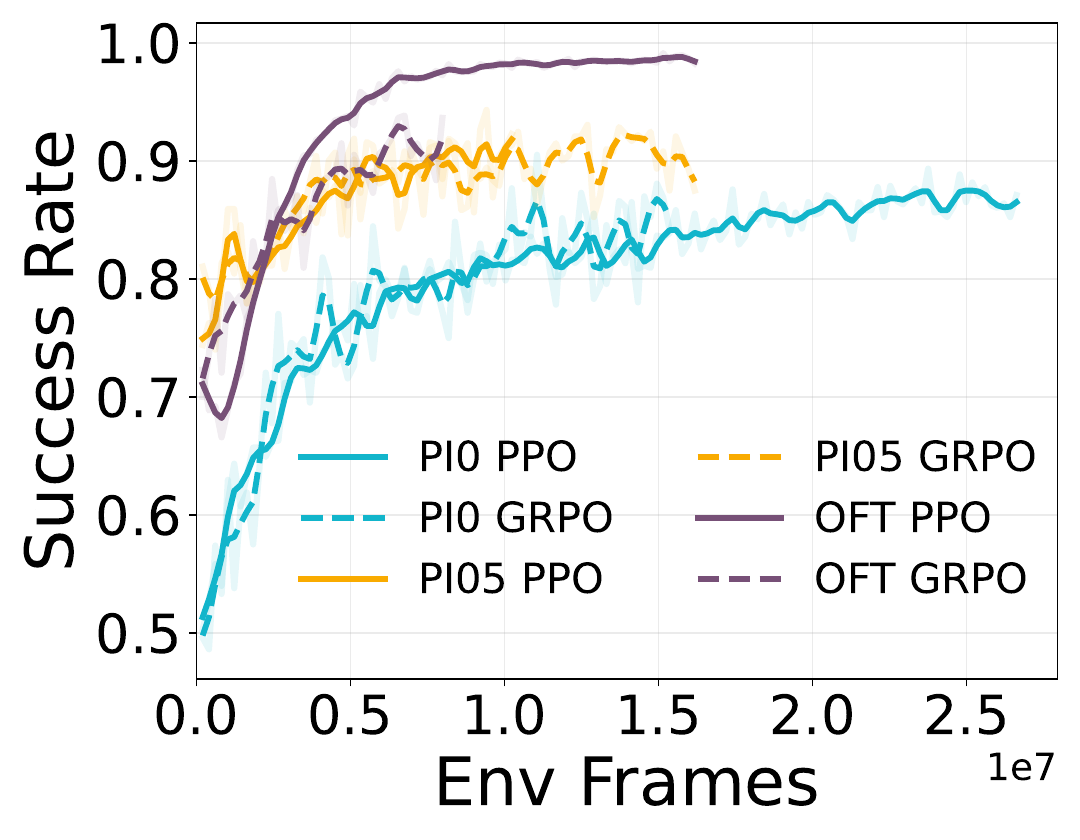}
    \caption{RoboTwin(Place Empty Cup)}
    \label{fig:robotwin-compare}
    \end{subfigure}
    \begin{subfigure}{0.3\linewidth}
    \includegraphics[width=\linewidth]{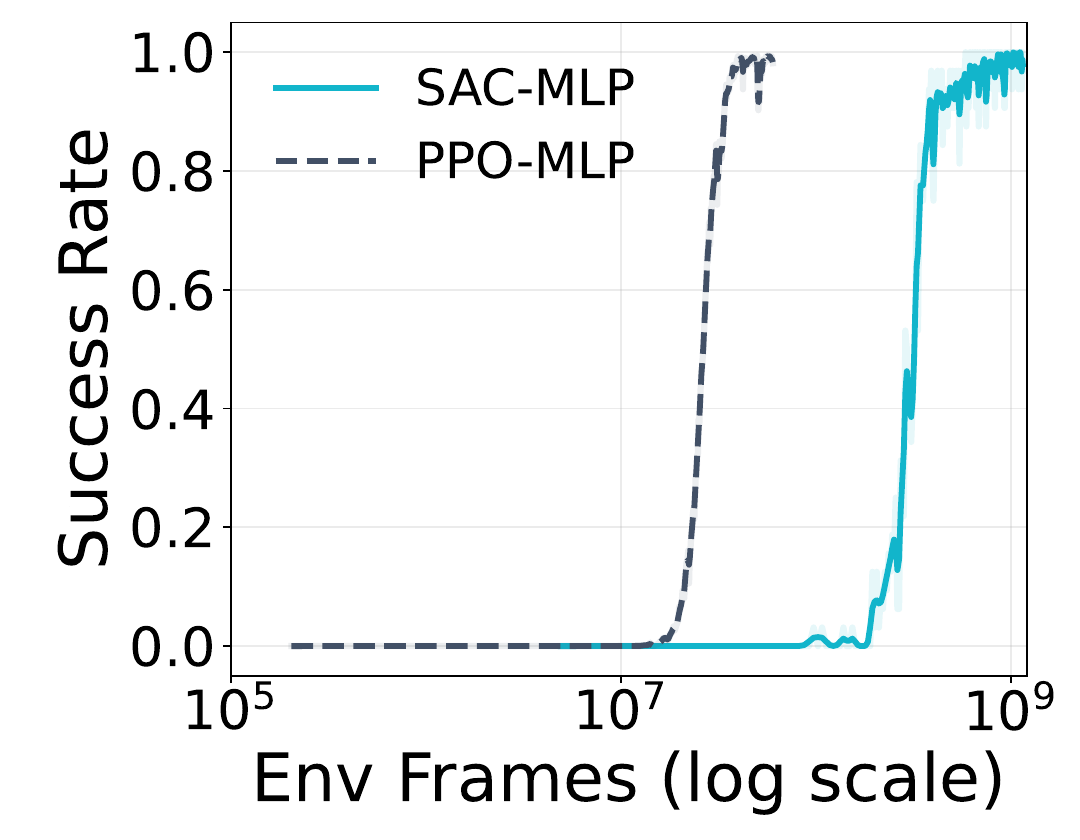}
    \caption{ManiSkill(PickCube), MLP}
    \label{fig:mlp}
    \end{subfigure}
    \begin{subfigure}{0.3\linewidth}
    \includegraphics[width=\linewidth]{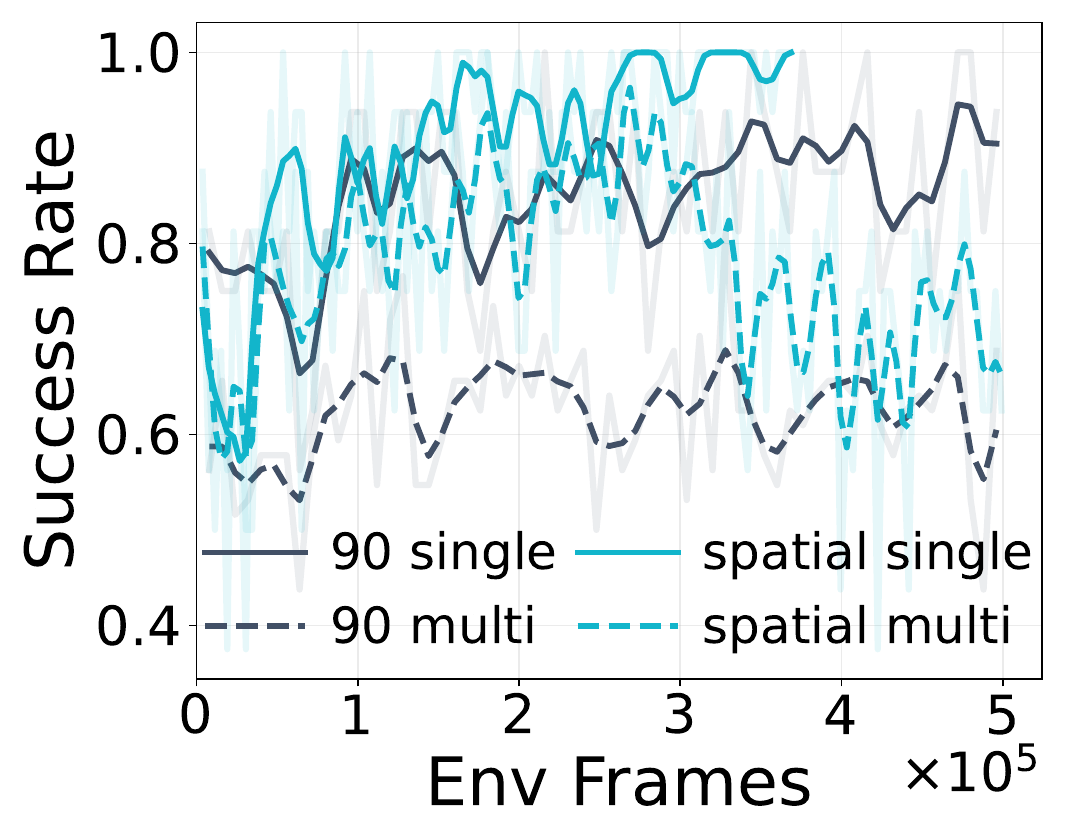}
    \caption{LIBERO, DSRL}
     \label{fig:dsrl}
    \end{subfigure}
    \caption{Extensibility of RLinf-VLA Across Models and Algorithms}
    \label{fig:extensible}
\end{figure}
\subsection{Extensibility of RLinf-VLA}
\label{app:extensible}
\subsubsection{Model Extensibility} RLinf-VLA also supports (a) flow-matching models and (b) small models. Figs~\ref{fig:robotwin-compare} and \ref{fig:mlp} present results on $\pi$-series models and MLPs, respectively. For $\pi$-series models, we adopt GRPO and PPO from \cite{chen2026pitextttrlonlinerlfinetuning}.

\subsubsection{Algorithm Extensibility} Our framework also supports off-policy methods such as SAC in multiple settings. However, applying SAC to VLA models remains an open problem, particularly in terms of achieving stable training. Nonetheless, the pipeline is fully functional. We demonstrate this capability using DSRL~\citep{wagenmaker2025steering} on $\pi_0$ (Fig.~\ref{fig:dsrl}) and SAC on MLP(Fig.~\ref{fig:mlp}). 

\subsection{Evaluation of OOD for ManiSkill}
\Cref{tab:mani-eval-ood-detailed} is the detailed results ManiSkill's experiments.
\begin{table}[h]
\centering
\caption{Detailed breakdown of Out-Of-Distribution evaluation results on ManiSkill. Values denote success rates (\%).}
\label{tab:mani-eval-ood-detailed}

    \begin{tabular}{lccc}
    \toprule
                    & \multicolumn{3}{c}{Out-Of-Distribution} \\
    Method          & Vision          & Semantic       & Execution      \\\midrule
    OpenVLA (Base)          & 38.75          & 35.94          & 42.11          \\
    OpenVLA (RLinf-GRPO)     & 74.69          & 72.99          & 77.86          \\
    OpenVLA (RLinf-PPO)      & 82.03          & \textbf{78.35} & \textbf{85.42} 
                            \\\midrule
    OpenVLA-OFT (Base)      & 27.73          & 12.95          & 11.72          \\
    OpenVLA-OFT (RLinf-GRPO) & 84.69          & 45.54          & 44.66          \\
    OpenVLA-OFT (RLinf-PPO)  & \textbf{92.11} & 64.84          & 73.57          \\\bottomrule
    \end{tabular}

\end{table}
\subsection{Ablation Studies}
\begin{figure}[htbp]
    \centering
    \begin{subfigure}{0.4\linewidth}
    \includegraphics[width=0.9\linewidth]{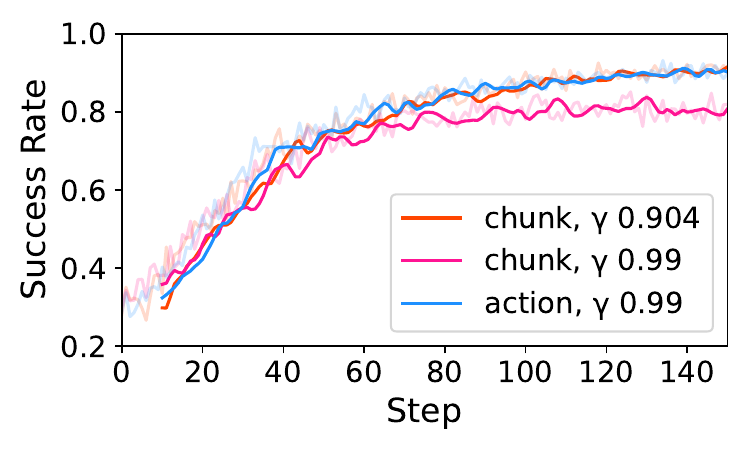}
    \caption{Success Rate}
    \label{fig:gamma-succ}
    \end{subfigure}
    \begin{subfigure}{0.4\linewidth}
    \includegraphics[width=0.9\linewidth]{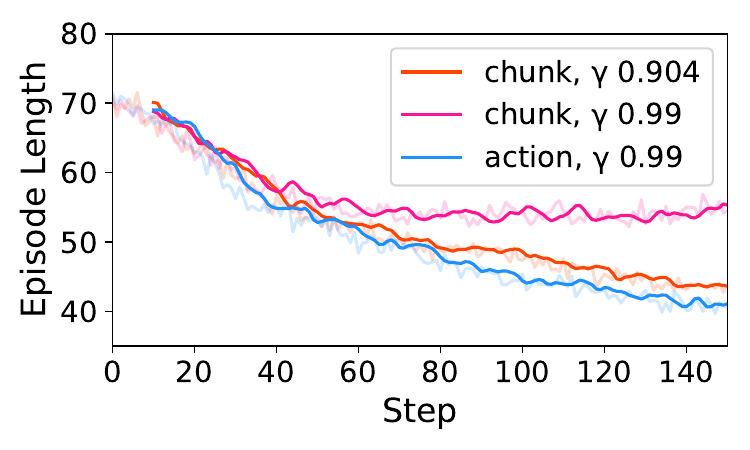}
    \caption{Episode Length}
    \label{fig:gamma-len}
    \end{subfigure}
    \caption{Ablation study of the discount factor $\gamma$ for action-level and chunk-level PPO training on ManiSkill with OpenVLA-OFT.}
    \label{fig:gamma}
\end{figure}

\paragraph{Gamma with action chunk}
At the chunk level, we treat an entire action chunk as a single RL action. This formulation requires careful selection of the discount factor $\gamma$. In our experiments, we observe that chunk-level training achieves performance comparable to action-level training only when the discount factors are properly aligned. Specifically, if the chunk size is $c$, the discount factors should satisfy
\[
\gamma_{\text{chunk}}
=
\gamma_{\text{action}}^{1/c}.
\]
The experimental results are shown in \Cref{fig:gamma}. In our setup, the chunk size is $c=8$. Accordingly, chunk-level training with $\gamma_{\text{chunk}}=0.904 \approx 0.99^{1/8}$ achieves performance comparable to action-level training with $\gamma_{\text{action}}=0.99$. In contrast, directly using $\gamma_{\text{chunk}}=0.99$ leads to substantially worse performance, highlighting the importance of properly rescaling the discount factor under chunk-level formulations.

Interestingly, this observation also highlights an important difference between embodied RL and RL for LLMs. In LLM settings, the discount factor is typically set to $\gamma=1$, since sequence generation naturally favors long-horizon credit assignment. In contrast, embodied RL often benefits from smaller discount factors. Intuitively, smaller $\gamma$ values encourage shorter-horizon behaviors and faster task completion, which are desirable in robotic manipulation tasks. As shown in \Cref{fig:gamma-len}, smaller $\gamma$ values lead to shorter trajectories on average. Moreover, when PPO with partial reset is enabled, shorter trajectories further improve data collection efficiency by allowing environments to reset and resume more frequently.

\begin{figure}[htbp]
    \centering
    \begin{subfigure}{0.3\linewidth}
    \includegraphics[width=0.9\linewidth]{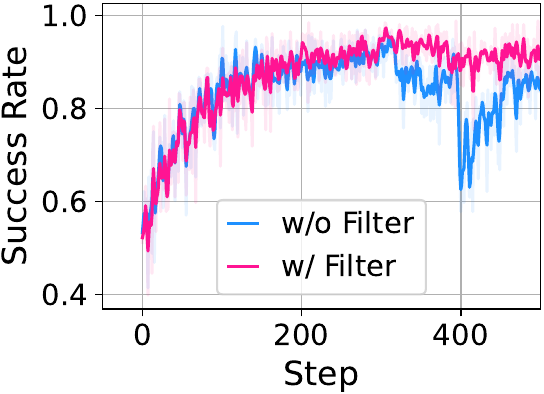}
    \caption{OpenVLA, ManiSkill}
    \label{fig:filter-openvla}
    \end{subfigure}
    \begin{subfigure}{0.3\linewidth}
    \includegraphics[width=0.9\linewidth]{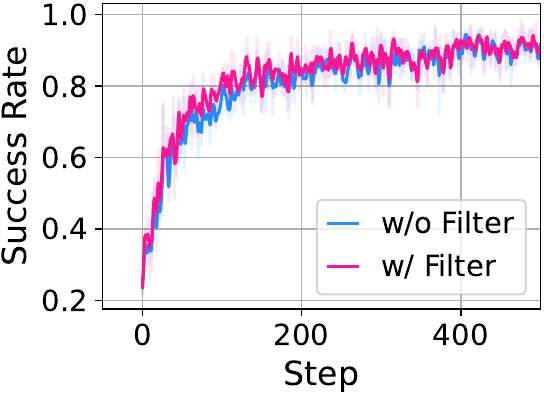}
    \caption{OpenVLA-OFT, ManiSkill}
    \label{fig:filter-oft-mani}
    \end{subfigure}
    \begin{subfigure}{0.3\linewidth}
    \includegraphics[width=0.9\linewidth]{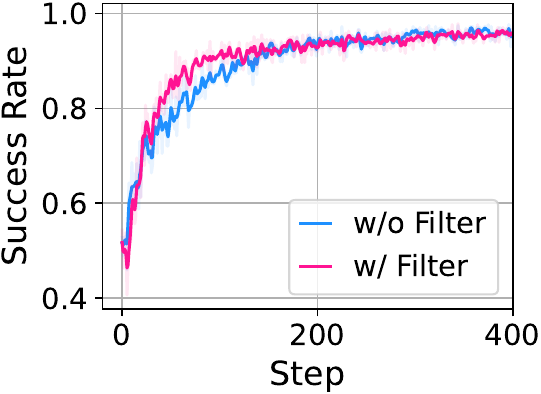}
    \caption{OpenVLA-OFT, LIBERO-Goal}
     \label{fig:filter-oft-goal}
    \end{subfigure}
    \caption{Ablation studies on success rate filtering under different settings.}
    \label{fig:filter}
\end{figure}

\paragraph{Success rate filtering}
As shown in \Cref{fig:filter}, we implement a success rate filter for GRPO that discards groups in which all trajectories have identical cumulative rewards. This mechanism can improve the stability of GRPO training. For instance, in the OpenVLA ManiSkill setting, training without the filter (``w/o Filter'') exhibits a clear collapse around step 400, whereas enabling the filter (``w/ Filter'') largely alleviates this issue. However, the benefit of the filter is not universal: in the OpenVLA-OFT ManiSkill setting and the OpenVLA-OFT LIBERO-Goal setting, its effectiveness is much less pronounced.

\begin{figure}[htbp]
    \centering
    \begin{subfigure}
    {0.4\linewidth}
        \centering
        \includegraphics[height=0.3in]{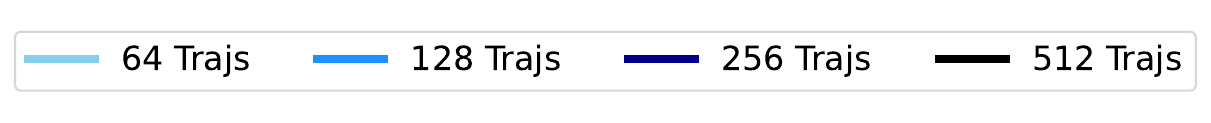} \\
        \vspace{-2mm}
        \includegraphics[width=0.8\linewidth]{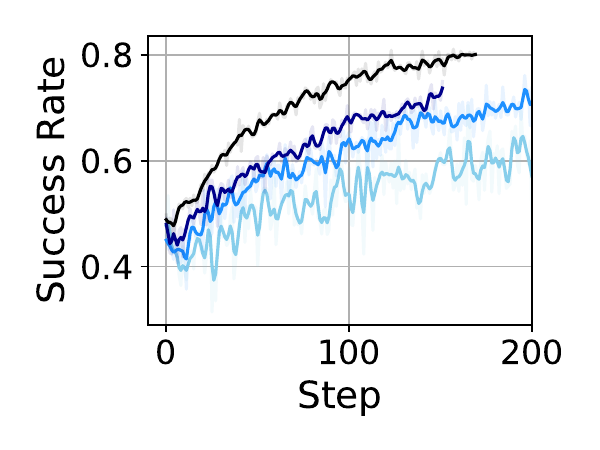}
        \vspace{-2mm}
        \caption{PPO, LIBERO-Long}
    \end{subfigure}
    \begin{subfigure}{0.4\linewidth}
        \centering
        \includegraphics[height=0.3in]{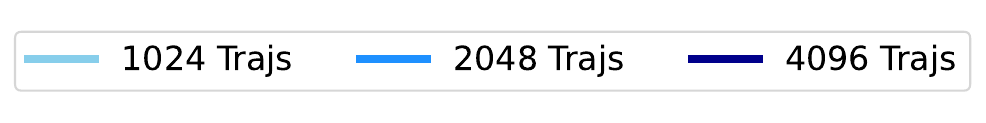} \\
        \vspace{-2mm}
        \includegraphics[width=0.8\linewidth]{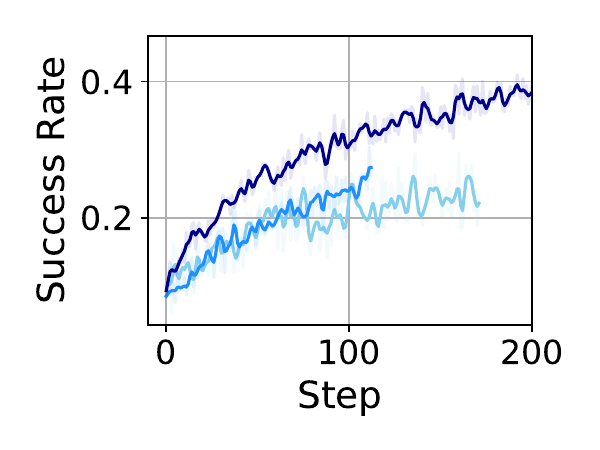}
        \vspace{-2mm}
        \caption{GRPO, LIBERO 130 tasks}
    \end{subfigure}

    \caption{Ablation study on rollout data size. Darker colors correspond to larger rollout datasets.}
    \label{fig:rollout}
\end{figure}

\paragraph{Effect of rollout data size}
The rollout batch size has a notable influence on RL performance, especially for on-policy algorithms such as PPO and GRPO. In general, larger rollouts per epoch enable more substantial policy improvement within each training iteration. As illustrated in \Cref{fig:rollout}, larger rollouts consistently achieve higher success rates when evaluated by training epochs.

\begin{figure}[htbp]
    \centering
    \begin{subfigure}{0.3\linewidth}
        \includegraphics[width=\linewidth]{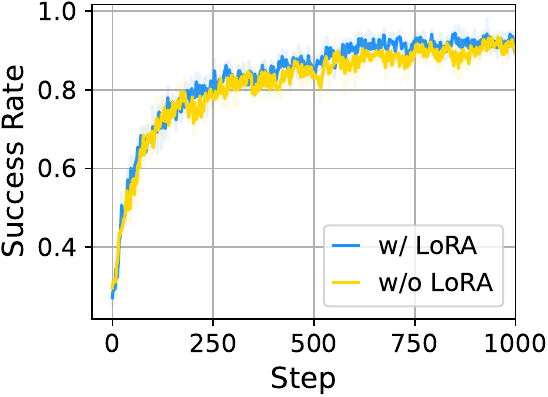}
        \caption{PPO}
        \label{fig:abl-lora-ppo}
    \end{subfigure}
    \begin{subfigure}{0.3\linewidth}
        \includegraphics[width=\linewidth]{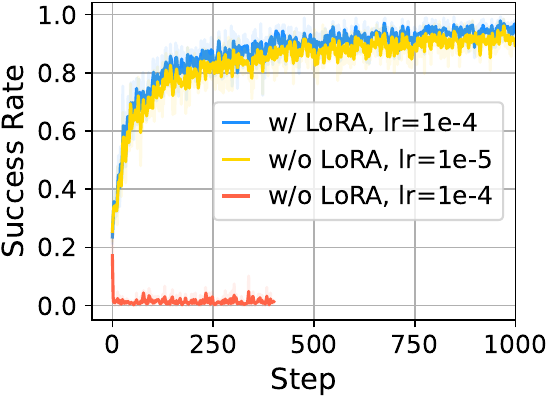}
        \caption{GRPO}
        \label{fig:abl-lora-grpo}
    \end{subfigure}
    \caption{Ablation of LoRA in ManiSkill with OpenVLA-OFT.}
    \label{fig:abl-lora}
\end{figure}

\paragraph{Using LoRA may not directly affect performance but often requires different hyperparameters}
\Cref{fig:abl-lora} shows the training curves with and without LoRA~\citep{lora}. The x-axis denotes the number of training epochs, while the y-axis shows the success rate in ManiSkill. The curves are overall similar, suggesting that LoRA itself does not substantially change performance. However, the choice of whether to use LoRA can influence the optimal hyperparameters. For example, in GRPO experiments (\Cref{fig:abl-lora-grpo}), using the same learning rate of $1 \times 10^{-4}$ leads to normal improvement with LoRA, but the success rate without LoRA collapses to zero. In contrast, when the learning rate is reduced to $1 \times 10^{-5}$, the non-LoRA setting also achieves stable improvement. These results indicate that different LoRA configurations may require separate hyperparameter tuning.

\end{document}